\documentclass{ecai} 

\usepackage{latexsym}
\usepackage{amssymb}
\usepackage{enumitem}
\usepackage{amsmath}
\usepackage{amsthm}
\usepackage{booktabs}
\usepackage{graphicx}
\usepackage{subfigure}
\usepackage{color}
\usepackage{hyperref}
\usepackage{multirow}
\usepackage{hhline}
\newcommand{\cN}{\mathcal{N}}
\usepackage{bbm}
\usepackage{bigstrut}
\usepackage{algpseudocode}
\usepackage[numbers]{natbib}
\usepackage{notoccite}

\usepackage[linesnumbered,ruled]{algorithm2e}

\if{0}
\usepackage{amsmath,amsfonts}
\usepackage{array}
\usepackage{textcomp}
\usepackage{stfloats}
\usepackage{url}
\usepackage{verbatim}
\usepackage{multirow}
\usepackage{hhline}
\usepackage{color}

\usepackage{mathtools}
\usepackage{microtype}
\usepackage{graphicx}
\usepackage{subfigure}
\usepackage{booktabs} 
\usepackage{bigstrut}
\usepackage{hyperref}
\usepackage{mathtools}
\usepackage{amsthm}
\usepackage{bbm}
\usepackage[capitalize,noabbrev]{cleveref}
\usepackage{algpseudocode}

\usepackage{xcolor}
\newcommand{\cN}{\mathcal{N}}
\fi

\newcommand{\BibTeX}{B\kern-.05em{\sc i\kern-.025em b}\kern-.08em\TeX}

\newcommand{\ouralgo}{DataDefense}

\begin{document}


\begin{frontmatter}


\paperid{1134} 


\title{A Data-Driven Defense against Edge-case Model Poisoning Attacks on Federated Learning}

\author[A]{\fnms{Kiran}~\snm{Purohit}}
\author[B]{\fnms{Soumi}~\snm{Das}}
\author[C]{\fnms{Sourangshu}~\snm{Bhattacharya}}
\author[D]{\fnms{Santu}~\snm{Rana}} 

\address[A, B, C]{Indian Institute of Technology, Kharagpur}
\address[D]{Deakin University, Australia}


\begin{abstract}
\color{red} \color{black}
Federated Learning systems are increasingly subjected to a multitude of model poisoning attacks from clients. Among these, edge-case attacks that target a small fraction of the input space are nearly impossible to detect using existing defenses, leading to a high attack success rate.
We propose an effective defense using an external \textit{defense dataset}, which provides information about the attack target. 
The defense dataset contains a mix of poisoned and clean examples, with only a few known to be clean. 
The proposed method, \textbf{\ouralgo{}}, uses this dataset to learn a \textit{poisoned data detector model} which marks each example in the defense dataset as poisoned or clean.
It also learns a \textit{client importance model} that estimates the probability of a client update being malicious. The global model is then updated as a weighted average of the client models' updates.
The poisoned data detector and the client importance model parameters are updated using an alternating minimization strategy over the Federated Learning rounds.
Extensive experiments on standard attack scenarios demonstrate that \ouralgo{} can defend against model poisoning attacks where other state-of-the-art defenses fail. In particular, \ouralgo{} is able to reduce the attack success rate by at least $\sim$ 40\% on standard attack setups and by more than 80\% on some setups.
Furthermore, \ouralgo{} requires very few defense examples (as few as five) to achieve a near-optimal reduction in attack success rate. 
\end{abstract}
\end{frontmatter}


\section{Introduction}

\label{sec:intro}

\color{black}

\textit{Federated Learning} (FL) \citep{mcmahan2017communication} enables multiple \textit{clients} to collaboratively build a \textit{global model} without compromising the privacy of their training data. However, the global model is vulnerable to {\it model poisoning attacks} \citep{bagdasaryan2020backdoor} where some clients may send malicious updates to the central server to compromise the global model.  
Untargeted model poisoning attacks, also known as Byzantine attacks \citep{blanchard2017machine, guerraoui2018hidden}, attempt to reduce the overall \textit{model accuracy} (MA). 
In this paper, we explore defenses against \textbf{targeted model poisoning attacks}, which are intended to achieve a specific objective.
One such class of targeted model poisoning attacks is \textit{backdoor attacks} \citep{bagdasaryan2020backdoor,harikumar2021scalable,wang2020attack}, which aim to achieve a particular malicious objective, such as misclassifying images of a particular class, like “airplane,” as “truck” \citep{bagdasaryan2020backdoor, li2022backdoor} at test time. 
The backdoor can be triggered by a pre-defined trigger patch in the test image \citep{harikumar2021scalable,li2022backdoor}, called the \textit{trigger-patch attack}.
Another special class of backdoor attacks are the \textit{edge-case attacks} \citep{wang2020attack}, which are proven to be hard to detect since they target only a small subset of the model's input space, which are the main subject of our study.

\begin{figure*}[!t]
\begin{center}
    \subfigure[]{\includegraphics[width = 2.0in]{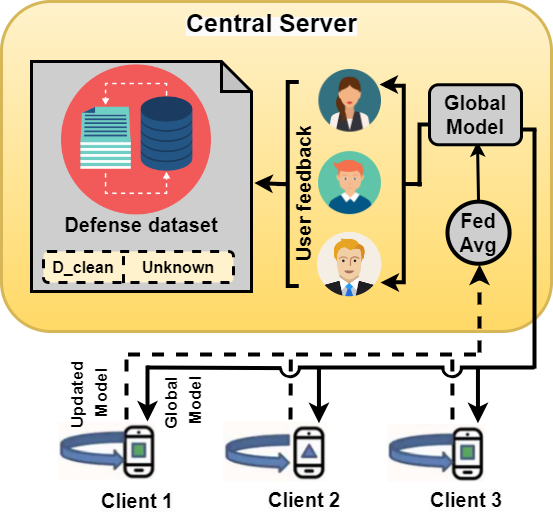}}
    \hspace{9mm}
	\subfigure[]{\includegraphics[width =4.6in]{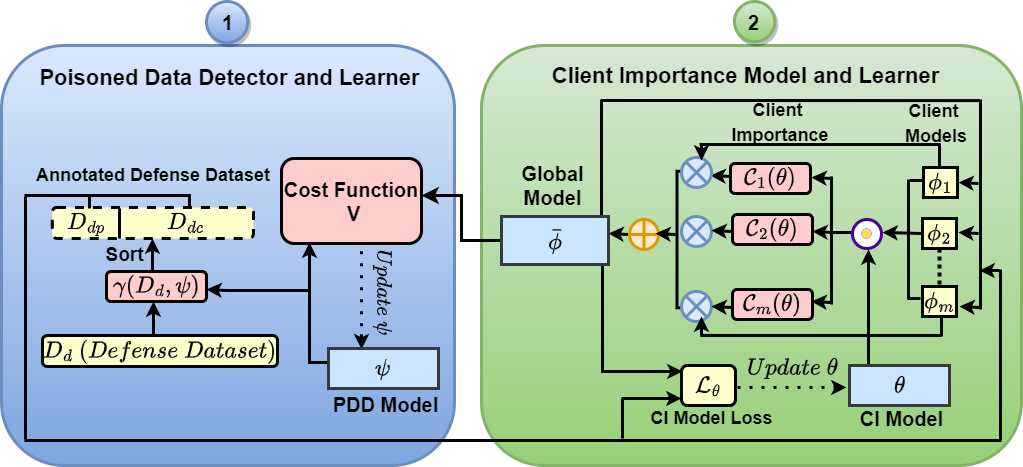}}
\end{center}
\caption{a) The overall scheme of \ouralgo{}. The defense dataset provides hints about possibly poisoned examples. b) Architecture Overview of the \ouralgo{}. Poisoned Data Detector (PDD) ($\gamma(\psi)$) detects the poisoned examples in the defense dataset, and the client importance model ($C(\phi,\theta)$) calculates the probability of each client
update being malicious.
}
\label{fig:flowchart}
\vspace{3mm}
\end{figure*}

\noindent\textbf{Threat model:} We base our threat model on two classes of attacks described in \citep{li2022backdoor}: (1) optimized attacks \citep{liu2018trojaning}, where the trigger patch is designed to maximize the attack success rate (ASR) with minimal impact on model accuracy, and (2) sample-specific \citep{nguyen2020input} attacks where the trigger pattern is specific to the sample. 
Both types of attacks can bypass Robust Aggregation-based defenses such as Robust Federated Aggregation (RFA) \citep{pillutla2019robust}, Norm Difference Clipping (NDC) \citep{sun2019can}, NDC-adaptive \citep{wang2020attack}, SparseFed \citep{panda2022sparsefed}, Krum, Multi-Krum \citep{blanchard2017machine}, and Bulyan \citep{guerraoui2018hidden}. A basic drawback of these defenses is their lack of knowledge about the attack target.
In many cases, however, the FL system administrator can anticipate such attacks and create synthetic poisoned examples.
Our main idea is to utilize a dataset of poisoned examples, called the \textbf{defense dataset}, to design a defense against these targeted attacks.
For instance, an attack on an FL-based object recognition model might aim to label an image of a particular politician as "Monkey" during an election. This may result in searches for the keyword ``Monkey'' showing images of that politician. \color{black}
The FL administrator could include such images in the defense dataset to pre-empt such attacks.
These possible poisoned examples can be created by a team of annotators from sources other than the FL clients, or generated using adversarial example generators \citep{yin2020dreaming}.
We assume that the clients do not have access to the defense dataset.
Figure \ref{fig:flowchart}-(a) illustrates our overall scheme. 
The defense dataset contains some marked clean examples (which are known to be not poisoned) and other unmarked examples that may or may not be poisoned. 
FLtrust \citep{cao2020fltrust}  proposes incorporating a ``root'' dataset of clean examples into the FL process to defend against Byzantine attacks. However, they do not target to defend against \textit{edge-case attacks}.

In this paper, we propose a novel method called \textit{\ouralgo{}}, for defending against targeted model poisoning attacks (Figure \ref{fig:flowchart}-(b)). 
Our defense works transparently with the federated averaging algorithm, where the central server updates the global model as a weighted average of the client models' updates. 
The client weights are computed using a \textit{client importance} model, that assigns high scores to benign clients and low scores to malicious clients.
To learn the client importance model, we utilize the \textit{defense dataset}, which is initially unmarked, with only a few examples known to be clean. Our method \ouralgo\ jointly learns the \textit{client importance} (CI) model and a \textit{poisoned data detector} (PDD) model, which marks the examples in the defense dataset as poisoned or clean. 
To learn the poisoned data detector, we introduce a novel \textit{cost function} that compares the markings given by a PDD network, with the loss incurred by the current global model, providing an aggregate cost of the PDD network.
We jointly minimize the cost function for the PDD model and the loss for the client importance model using an alternative minimization strategy.\\

\noindent\textbf{Contributions.} The key contributions of our work are:
\begin{itemize}
    \item We propose a novel approach \textit{\ouralgo{}}, to detect and defend against \textit{edge-case} (hard-to-detect) backdoors in FL.

    \item We propose an alternating minimization-based approach to jointly learn the \textit{client importance} and \textit{poisoned data detector} models.

    \item  Our experiments demonstrate that \ouralgo\ outperforms baselines with a maximum of $\sim 15\%$ attack success rate (ASR), as opposed to a minimum of $\sim 27\% $ ASR (Table \ref{tab:comparison}).

    \item We show that PDD can detect poisoned data points correctly (Figure \ref{fig:theta_gamma_fig}-a), and client importances are learned correctly (Figure \ref{fig:theta_gamma_fig}-b). We also show that stopping the updates of both PDD and CI models makes the performance of the global model worse (Figure  \ref{fig:trigger_14}).
    
\end{itemize}

\color{black}



\subsection{Related Work}
\label{sec:related}

A recent survey about attacks and defenses on machine learning (ML) models \citep{tian2022comprehensive} categorizes the attacks as targeted (that target a small subspace of the input examples) and untargeted attacks (that target the entire learning process). ML models are also susceptible to adversarial
attacks \citep{zheng2022data}. The pitfalls of mentioned above can be addressed by model compression \citep{zheng2022data, 10.1145/3564121.3564139}. In this work, we focus on making a defense against targeted attacks like edge case \citep{wang2020attack} or trigger patch attacks \citep{gu2017badnets, harikumar2021scalable}.
One such class of targeted attacks are backdoors \citep{chen2017targeted, saha2020hidden} which infect the ML model, and are detected only when an appropriate attack is triggered. 
Federated Learning (FL) systems are shown to be particularly vulnerable to backdoors because of the accessibility of the global model to the participating clients \citep{bagdasaryan2020backdoor}.
Wang et al. \citep{wang2020attack} showed that backdoors which affect a small subspace of the input samples, called \textit{edge case attacks}, can be hard to defend against. To the best of our knowledge, most defenses against such attacks are ineffective in reducing the ASR significantly.

\textbf{Defense Techniques:}
Defenses against federated backdoors can be broadly categorized into three main groups: model refinement, robust aggregation, and certified robustness. \textit{Model refinement} defenses aim to eliminate potential backdoors from the global model through techniques like fine-tuning \citep{wu2020mitigating} or distillation and pruning \citep{li2020neural,lin2020ensemble,liu2018fine}. 
In contrast, \textit{robust aggregation} defenses \citep{awan2021contra,blanchard2017machine,cao2020fltrust,chen2021fedequal,fung2020limitations,guerraoui2018hidden,ozdayi2021defending,yin2018byzantine} identify and exclude malicious weights and gradients from suspicious clients using anomaly detection or dynamically adjusting the importance of clients. 
\textit{Certified robustness} \citep{cao2021provably,cao2022flcert,xie2021crfl} focuses on providing guarantees that the model's predictions remain unchanged even if certain features in the local training data of malicious clients are modified within certain constraints.

Our robust aggregator \textit{\ouralgo{}}, being in the federated averaging setup, does not have any gradient information and thus is different from methods like SparseFed \citep{panda2022sparsefed}, FedEqual \citep{chen2021fedequal} and CONTRA \citep{awan2021contra}. 

\section{\ouralgo{} against Targeted Attacks on FL}
\label{sec:method}

\newcommand{\cC}{\mathcal{C}}

The main objective behind the proposed \ouralgo\ framework is to use a \textit{defense dataset} to provide hints towards potential targeted model poisoning attacks. 
In this section, we describe the architecture behind various components of the proposed \ouralgo{} framework for defending against attacks in the Federated Learning (FL) setting.
The main challenge behind incorporating the defense dataset is to automatically infer the poisoned examples, which ultimately provides information to the central server for learning the robust global model. We begin by formalizing the Federated Learning setting and describe the defense framework (Section \ref{sec:framework}).
Then we describe the algorithm for learning the poisoned data detector (PDD) for inferring poisoned examples in the defense dataset (Section \ref{sec:learngamma}). Finally, we describe the algorithm for jointly learning the poisoned data detector and the client importance model (Section \ref{sec:finalalgo}).

\newcommand{\cX}{\mathcal{X}}
\newcommand{\cY}{\mathcal{Y}}

\subsection{Framework for \ouralgo{}}
\label{sec:framework}


We follow the federated averaging framework for learning the model \citep{mcmahan2017communication}. Let $\bar{\phi}^t$ denote the global model computed by the central server at time $t$, and $\phi_j^t, j=1,...,M$ denote the local models for each client $j$, $n_j$ is the number of data samples held by the client $j$ at time $t$ and $n$ is the total number of data samples among all $M$ clients (that is: $\sum_{j=1}^M n_j$). The update equations can be written as: 
\begin{align}
    \mbox{Global model update: }& \bar{\phi}^t = \bar{\phi}^{t-1} + \sum_{j=1}^{M} \frac{n_j}{n} (\phi_j^t - \bar{\phi}^{t-1})  \label{eq:fedavg}  \\
    \mbox{Client model update: }& \phi_j^t = \phi_j^{t-1} - \eta \nabla l(\bar{\phi}^{t-1}); \ \forall j=1,...,M 
    \nonumber
\end{align}

\textbf{Defense Architecture:}
The main idea behind the proposed \ouralgo, is to use a \textit{defense dataset} comprising of some examples that are possible targets of the model poisoning attack, to learn a better defense mechanism over time. 
The defense dataset, $D_d$, consists of unmarked poisoned and clean examples, with only a few examples marked as clean. These examples can be collected from unreliable sources, e.g. user feedback.
The design of \ouralgo\ has two main components (see Figure \ref{fig:flowchart}-(b)):
\begin{itemize}
    \item \textbf{Poisoned Data Detector Learner:} (described in Section \ref{sec:learngamma}) This learns a scoring function $\gamma(x,y;\psi)$ for a datapoint $(x,y)\in D_d$, which reflects the probability of the datapoint being poisoned. $\psi$ denotes the parameters for the poisoned data detector model. This model is used to mark the data points in the defense dataset for learning the client importance model.
    
    \item \textbf{Client Importance Model Learner:} This uses a marked defense dataset $D_d$, (each datapoint is marked as poisoned or clean by the poisoned data detector model) to learn importance scores for the client model updates. $\cC(\phi^t_j,\theta)$ denotes the client importance for client $j$ (with local model update $\phi^t_j$. $\theta$ denotes the parameters of the \textit{client importance model}.

\end{itemize}

\color{black}

\ouralgo\ works by weighing down malicious client models with \textit{client importance weights}, $\cC(\phi_j^t,\theta)$ for client $j$. The \textbf{revised global model update equation} (Equation \ref{eq:fedavg}) becomes:
\begin{equation}
    \bar{\phi}^t(\theta) = \bar{\phi}^{t-1}(\theta) + \sum_{j=1}^{M}\mathcal{C}(\phi_j^t, \theta)   (\phi_j^t - \bar{\phi}^{t-1}(\theta))
    \label{eq:global}
\end{equation}

Here, we ensure that the client importance model forms a discrete probability distribution over the clients, i.e. $\sum_{j=1}^M \cC(\phi_j,\theta) = 1$ and $\cC(\phi_j,\theta)\geq 0$.
 Note that the client importance function $\cC$ depends on the parameters of the client model $\phi_j$ but not on the client identity $j$. Hence, our defense is resilient to malicious clients changing their identities or operating with a small number of different identities.

Assuming that the defense dataset $D_d$ is partitioned into poisoned $D_{dp}$ and clean  $D_{dc}$ datasets by the PDD model, we use three features for estimating client importance $\cC$:
\begin{itemize}
\item Average cross-entropy loss of the client model on the clean defense dataset,\\
$\bar{L}_{dc}(\phi_j) = \frac{1}{|D_{dc}|} \sum_{(x,y)\in D_{dc}} l(x,y;\phi_j)$. 
\item Average cross-entropy loss of the client model on the poisoned defense dataset,\\ $\bar{L}_{dp}(\phi_j)= \frac{1}{|D_{dp}|} \sum_{(x,y)\in D_{dp}} l(x,y;\phi_j)$).
\item $L_2$-distance of the client model from the current global model: \\$dist(\phi_j) = \|\phi_j - \bar{\phi} \|_2$.
\end{itemize}

The first two features describe the performance of the client model w.r.t. the current marked defense dataset, while the third feature provides a metric for deviation of the client model from the global model for performing robust updates. Given the input feature vector $s(\phi_j) = [\bar{L}_{dc}(\phi_j), \bar{L}_{dp}(\phi_j), dist(\phi_j) ]$.
The client importance for each client $j$ is calculated as a single feedforward layer with \textit{ReLu} activation followed by normalization across clients:
\begin{equation}
    \cC(\phi_j;\theta) = \frac{ReLu(\theta^T s_j)}{\sum_{j=1}^{M}ReLu(\theta^T s_j)} \label{eq:clientimportance}
\end{equation}

The above Equation \ref{eq:clientimportance} for client importance along with Equation \ref{eq:global} define our defense mechanism, given a learned parameter $\theta$. 

In order to learn the parameter $\theta$ from the split defense dataset $D_{d}=D_{dp}\cup D_{dc}$, we define two loss functions: (1) \textit{clean loss}, $l_c(.)$, and (2) \textit{poison loss}, $l_p(.)$. 
Given the predictive model $f(y|x,\phi)$, the clean loss encourages the correct classification of a clean datapoint $(x,y)$. Hence, we define it as: 
\color{black}
\begin{equation}
l_c((x,y);\phi) = - \log( f(y|x,\phi) )
\end{equation}

On the contrary, for a poisoned input datapoint $(x,y)$, the probability for class $y$ predicted by the model $f(y|x,\phi)$, should be low. Hence, we define the poison loss function as:
\begin{equation}
l_p((x,y);\phi) = - \log( 1 - f(y|x,\phi))
\end{equation}

Combining the above losses $l_c$ and $l_p$ on the corresponding defense datasets, we define the combined loss function described below. We learn the parameter $\theta$ for an optimal defense by minimizing the below loss function. 
\begin{equation}
\mathcal{L}_{\theta}(\theta| D_{dc}, D_{dp}) = \hspace{-4mm} \sum_{(x,y) \in D_{dc}} \hspace{-4mm}l_c ((x,y); \bar{\phi}(\theta)) +\hspace{-4mm} \sum_{(x,y)\in D_{dp}}\hspace{-4mm} l_p((x,y); \bar{\phi}(\theta))
\label{eq:updatetheta}
\end{equation}



\subsection{Learning Poisoned Data Detector}
\label{sec:learngamma}

The main motivation behind the poisoned data detector (PDD) is to classify data points in the defense dataset as poisoned or clean. 
We intend to learn a neural network $\gamma((x,y);\psi)$ for estimating a score of an input datapoint $(x,y)$ being poisoned, where $\psi$ denotes the parameters of the neural network. 
We treat $\gamma$ as a ranking function and mark the top-$\beta$ fraction of $D_d$ as poisoned (belonging to $D_{dp}$).
The parameter $\psi$ of the PDD function is then trained using a novel cost function. 
We note that this design also introduces the user-controllable hyperparameter $\beta$ which can be used to specify the expected number of poisoned datapoints in $D_d$ depending on the current attack and total number of datapoints in $D_d$.
If new data points are added to the defense dataset $D_d$ to incorporate new information about the attack during the FL rounds, then $\beta$ needs to be tuned to reflect the changed fraction of poisoned data points.


\begin{algorithm}[t]
    \SetAlgoLined
    \small
  \caption{\ouralgo}
  \label{algo:learning}
  \small
    	\textbf{Input:} Fraction of poisoned samples- $\beta$, Defense Dataset-     $D_d$, Clean samples- $D_{clean} \subseteq D_d$\\

            \textbf{Initialize:} \\ 
            
            \hspace{2mm} $\psi^0$ consistent with $D_{clean}$ \Comment{Eq \ref{eq:psi-consistent}}\\

            \hspace{2mm} $\theta^0 \leftarrow \cN(0,1)$ \Comment{Sampling from a gaussian }\\
            
    \For {$t=1$ {\bfseries to} $T$ }{
		  $D^t_{dp} \leftarrow$ top $\beta$ fraction of $D_d$ using $\gamma(., \psi^{t-1} )$
    
            $D^t_{dc} \leftarrow$ rest of $D_d$ using $\gamma(., \psi^{t-1} )$

            $\psi^t \leftarrow  \min_{\psi}\mathcal{L}(\psi^{t-1},\theta^{t-1}|D_d,D_{clean},\bar{\phi}^{t-1})$ \Comment{Eq \ref{eq:learnpsi}}

            Receive updates $\{\phi_j^t\}^{M}_{j=1}$ from clients 
            
            Calculate client importance $\cC(\phi_j^t;\theta^{t-1})$ \Comment{Eq \ref{eq:clientimportance}}
            
            Calculate new global model $\bar{\phi}^t(\theta^{t-1})$ \Comment{Eq \ref{eq:global}}

            $\theta^t \leftarrow   \min_{\theta}\mathcal{L}(\theta^{t-1}|D^t_{dc},D^t_{dp})$ \Comment{Eq \ref{eq:updatetheta}}
            
            }
     	\textbf{Output:  } Global Model $\bar{\phi}$
\end{algorithm}

\textbf{Architecture of PDD Network: }
The architecture of the neural network for PDD depends on the end task. However, we follow a general principle where we simulate the process of task prediction followed by comparison with the ground truth label.
In this paper, we describe a network for the multi-class classification task.
As a part of simulating the task predictions, the input features $x$ of every datapoint are passed through a feature encoder, $FE(x)$ which is related to the task to produce an intermediate embedding $h_1$. We assume that this feature encoder cannot be corrupted by the attack on the FL system.
Also, the parameters of the feature encoder are kept fixed to learn the parameters of the PDD network.
The intermediate embedding $h_1$ is then passed through a shallow multi-layer perceptron (MLP) to output the logit vector of the predicted output $\hat{y}$.
A concatenation of the predicted output $\hat{y}$, and the one-hot encoding of true label $y$ are then passed through a 2-layer MLP network, followed by a normalization (described later) to produce the final output $\gamma(x)$.
The following equations describe the architecture:

\begin{align}
h_1(x) = FE(x); 
&& h_2(x|\psi) = ReLU(W_1 h_1(x))\nonumber \\
 \hat{y}(x|\psi) = Soft( W_2 h_2(x)); 
&& g_1(x,y|\psi) = ReLU(W_3 [\hat{y}(x), y])\nonumber\\
g_2(x,y|\psi) = W_4 g_1(x,y); 
&&  \boldsymbol{\gamma((x,y)|\psi)} = Norm(g_2(x,y),D_d) 
\label{eq:gamma}
\end{align}

Hence, the parameter set for the PDD network $\gamma(x,y;\psi)\in [0,1]$ is: $\psi = [ W_1, W_2, W_3, W_4 ] $.
We note that $\gamma(x,y)$ for $(x,y)\in D_d$ is designed to model the probability of $(x,y)$ being a poisoned datapoint.
The output of the score predictor MLP $g_2(x,y|\psi)$, is passed through a normalization layer -  $Norm(g_2(x,y), D_d)$, which depends on the output of the entire defense dataset $D_d$. Many output normalization techniques have been studied in the neural network literature, with the most prominent being Batch Normalization \citep{ bjorck2018understanding,ioffe2015batch} and many other enhancements and variants (see e.g. \citep{awais2020revisiting,lubana2021beyond}). While Batch normalization (BN) is used to speed up training and reduce internal covariate shift, it is not known to induce a discrete distribution over the dataset (batch in this case). The main problem is that the output of BN can also have negative values, and hence cannot be interpreted as probability values. Awais et al. \citep{awais2020revisiting} report that MinMax normalization also performs similarly to Batch normalization while achieving a discrete distribution over the defense dataset. Hence we use MinMax normalization to calculate the final score $\gamma$, as follows: 
\begin{align}
    min = \min_{(x_i,y_i)\in D_d} g_2(x_i,y_i);
     \hspace{5mm} max = \max_{(x_i,y_i)\in D_d} g_2(x_i,y_i) \nonumber \\
     Norm(g_2((x,y),D_d)) =
    \frac{(g_2(x,y) - min)}{(max - min)},\ \forall (x,y)\in D_d
\end{align}


\color{black}

We note that the output of the $Norm$ layer is between $0$ and $1$ and adds up to a number less than $|D_d|$. While we could have normalized the summation to be equal to $1$, it would have made the scoring function output scale down with the increase in $|D_d|$. Hence, we use the above-defined form of normalization for the $\gamma$ function used for ranking the examples in the defense dataset according to their probability of being poisoned example.

\begin{table*}[t]
\scriptsize
\caption{ Comparing the model accuracy (MA) and attack success rate (ASR)
of various defenses on different datasets under PGD with model replacement attack after 1500
FL iterations.}
\label{tab:comparison}
\begin{center}
\begin{small}
\resizebox{\textwidth}{!}{%
\begin{tabular}{|l| c c| c c| c c|c c|c c| c c|}
\hline\

                            \multirow{3}{*}{\textbf{Defenses}}
                           & \multicolumn{2}{c}{\begin{tabular}[c]{@{}c@{}}\textbf{CIFAR-10}\\ \textbf{Label Flip}\end{tabular}}
                           & \multicolumn{2}{|c}{\begin{tabular}[c]{@{}c@{}}\textbf{CIFAR-10}\\ \textbf{Southwest}\end{tabular}} & \multicolumn{2}{|c}{\begin{tabular}[c]{@{}c@{}}\textbf{CIFAR-10}\\\textbf{Trigger Patch}\end{tabular}}
                            & \multicolumn{2}{|c}{\begin{tabular}[c]{@{}c@{}}\textbf{CIFAR-100}\\\textbf{Trigger Patch}\end{tabular}}
                           & \multicolumn{2}{|c}{\textbf{EMNIST}} 
                           & \multicolumn{2}{|c|}{\textbf{Sentiment}}

                           \\\cline{2-13}   
                           
& MA(\%)                                            & ASR(\%) 
                    & MA(\%)                                            & ASR(\%) 
                        & MA(\%)                                              & ASR(\%)    & MA(\%)       & ASR(\%) 
                    & MA(\%)                                            & ASR(\%) 
                        & MA(\%)                                              & ASR(\%)                                             \\\hhline{=============}
No Defense &86.38 &27.56 &86.02  &65.82  &86.07  &97.45 &63.55 & 100.00 &99.39  &93.00  &80.00   &100.0 \\
Krum 
&73.67 &4.10   & 82.34  & 59.69  & 81.36  & 100.00  &62.63 & 95.00  & 96.52  & 33.00  & 79.70   & 38.33  \\
Multi-Krum 
 &83.88 &2.51 & 84.47  & 56.63  & 84.45   & 76.44   &63.46 & 65.00 & 99.13  & 30.00  & 80.00  & 100.0  \\
Bulyan 
 &83.78 &2.34 & 84.48  & 60.20 & 84.46  & 100.00  &63.40 & 75.00    & 99.12  & 93.00  & 79.58  & 30.08   \\
Trimmed Mean 
&83.98 &2.45  & 84.42  & 63.23 & 84.43  & 44.39   &63.35 &70.00   & 98.82  & 27.00   & 81.17  & 100.0  \\
Median 
 &64.67 &5.87 & 62.40  & 37.35 & 62.16  & 31.03   &42.78 &20.54   & 95.78  & 21.00   & 78.52  & 99.16   \\
RFA 
&84.59 &15.78 &84.48  &60.20  &84.46  &97.45  &62.70 &100.00 &99.34  &23.00  &80.58  &100.0\\
NDC 
&84.56 &10.34 &84.37  &64.29  &84.44  &97.45  &62.90 &100.00  &99.36  &93.00  &80.88  &100.0 \\
NDC adaptive 
&84.45   &9.78  &84.29  &62.76  &84.42  &96.43   &62.78 &95.00  &99.36  &87.00  &80.45  &99.12 \\
Sparsefed 
 &81.35 &6.45 & 84.12  & 27.89  & 84.38  &11.67  &61.23 &20.36  & 99.28 & 13.28  & 79.95  & 29.56    \\
\textbf{\ouralgo{}}     & \textbf{84.73}   &  \textbf{0.01}   & \textbf{84.49}      &  \textbf{15.30}         & \textbf{84.47}   &  \textbf{2.04}   & \textbf{63.53} &  \textbf{8.34}                            & \textbf{99.37}                  
& \textbf{4.00}    & \textbf{81.34}  &  \textbf{3.87}   \\\hline
\end{tabular}
}
\end{small}
\end{center}
\end{table*}

\textbf{Loss function for learning PDD:}
Neural ranking loss functions can be broadly categorized into Pointwise, Pairwise, and Listwise \citep{guo2020deep}. In this paper, we design a pointwise loss function.
To estimate the “correctness" of ranking of defense datapoints induced by a PDD network $\gamma(.,\psi)$, we define \textit{cost function} $V$, which should be lower for a correct ordering, compared to an incorrect ordering.
For a given datapoint $(x,y)$, the key idea behind designing the cost function $V$ is to postulate a relationship between the score $\gamma((x,y);\psi)$ and the difference between poisoned and clean losses $(l_{p}((x,y);\bar{\phi}) - l_{c}((x,y);\bar{\phi}) )$, given a global model ($\bar{\phi}$).
Recall that the clean loss $l_c(x,y)$ is expected to be low for a clean datapoint $(x,y)$ and poisoned loss $l_{p}(x,y)$ is expected to be low for a poisoned datapoint.
Assuming that the current global model $\bar{\phi}$ is correct, we make the following observations:
\begin{itemize}
\item When poisoned loss $l_p$ is low, and hence $(l_{p}((x,y);\bar{\phi}) - l_{c}((x,y);\bar{\phi}) )$ is negative, $(x,y)$ is a poisoned datapoint and hence the corresponding $\gamma((x,y);\psi)$ should be high. Hence, the product $\gamma((x,y);\psi) (l_{p}((x,y);\bar{\phi}) - l_{c}((x,y);\bar{\phi}) )$ should be high negative number, or low.
\item When clean loss $l_c$ is low, and hence $(l_{p}((x,y);\bar{\phi}) - l_{c}((x,y);\bar{\phi}) )$ is positive, $(x,y)$ is a clean datapoint and hence the corresponding $\gamma((x,y);\psi)$ should be low. Hence, the product $\gamma((x,y);\psi) (l_{p}((x,y);\bar{\phi}) - l_{c}((x,y);\bar{\phi}) )$ should be a low positive number.
\end{itemize}

Combining the observations, we propose following \textbf{cost function}:
\begin{equation}
    V(\psi|D_d,\bar{\phi}(\theta)   ) = \hspace{-4.5mm}\sum_{(x,y)\in D_d} \hspace{-3mm} \gamma((x,y);\psi) (l_{p}((x,y);\bar{\phi}) - l_{c}((x,y);\bar{\phi}) )
    \label{eq:Value}  
\end{equation}

Note that the minimization of the above cost function for learning $\gamma(.,\psi)$ with only positivity constraints will result in a trivial solution of $\gamma((x,y);\psi) = 0$ for all $(x,y)$. 
However, the normalization step in the last layer of the $\gamma$ network ensures that the maximum value of $\gamma$ is $1$ for some datapoint in the defense dataset, and the minimum value is $0$. 
Hence, as long as the defense dataset has at least one poisoned datapoint and at least one clean datapoint, we are ensured of a good spread of $\gamma$-values.



Our architecture for the $\gamma$ network relies on the fact that $\hat{y}$ truly predicts the softmax probabilities for the end task. To ensure this, we use a small dataset of clean datapoints $D_{clean}$ to add the cross-entropy loss on predicted $\hat{y}$ as:
$ L_{pred}(\psi,D_{clean}) = \sum_{(x,y)\in D_{clean}} L_{CE}(\hat{y}(x|\psi),y) $.
The objective function for learning PDD parameters $\psi$ is given by:
\begin{equation}
\mathcal{L}_{\psi}(\psi,\theta|D_d,D_{clean},\bar{\phi}) = V(\psi|D_d,\bar{\phi}(\theta)) + \lambda L_{pred}(\psi,D_{clean}) \label{eq:learnpsi}
\end{equation}

Here, $\lambda$ is a tradeoff parameter for assigning importance to the correct prediction of $\hat{y}$ and a low value of the cost function. Higher $\lambda$ assigns high importance to the accuracy of $\hat{y}(x)$. It can be chosen using trial and error using the accuracy of $\hat{y}(x)$ on $D_{clean}$ which should be comparable to that of the global model $\bar{\phi}$.


\subsection{Joint Learning of Client Importance and PDD}
\label{sec:finalalgo}

The overall objective of the \ouralgo\ algorithm is to learn the parameters $\theta$ and $\psi$ for the client importance and PDD models, respectively. This is a coupled optimization problem since ${L}_{\theta}$ depends on $\psi$ through $D_{dc}, D_{dp}$, and ${L}_{\psi}$ depends on $\theta$ through $\bar{\phi}$. Algorithm \ref{algo:learning} describes our method for estimating $\theta^*$ and $\psi^*$ on the central server.
We use alternating minimization updates for both $\theta$ (line 12) and $\psi$ (line 8), along with the federated learning rounds for learning $\bar{\phi}$, to achieve the above objective. 
Line 3 initializes $\psi^0$ using an initialization that is consistent with $D_{clean}$ as described below. Line 4 sets up an initial $\theta^0$ randomly. The idea is to have a random initialization of $\cC(\phi_j)$ in the absence of any information about the client's importance. Alternatively, one could also initialize $\cC(\phi_j)$ uniformly randomly. Lines 5--12 show the FL rounds and the steps taken by the central server to aggregate the client models $\phi^t_j, j=1,...,M$ for defending against malicious clients. Lines 6 and 7 calculate the poisoned and clean examples $D_{dp}$ and $D_{dc}$ respectively using $\gamma$. Line 8 updates the PDD model parameter $\psi$ using the combined loss calculated from $D_{dp}$, $D_{dc}$, and $D_{clean}$. Note that the examples of $D_{clean}$ which is also part of $D_d$ can be part of $D_{dp}$ or $D_{dc}$. However, we treat them as part of $D_{clean}$. After receiving the client models in line 9, we calculate the client importance (line 10) and then calculate a new global model (line 11) which is then used in line 12 for updating the parameters of the client importance model, $\theta$.
Note that, this algorithm is a special case of weighted federated averaging which is known to converge when the weight distribution is fixed, (see e.g. \citep{li2020convergence}). Hence, for attacks where the client importance for each client is converged, Algorithm \ref{algo:learning} will also result in a convergent $\bar{\phi}^t$.

\textbf{Consistent Initialization of $\mathbf{\psi^0}$ with $\mathbf{D_{clean}}$: }
In line 3 of Algorithm \ref{algo:learning}, the objective is to provide a reasonable initial estimate for $\psi^0$ such that $\gamma(.,\psi^0)$ is consistent with $D_{clean}$. 
We formulate this problem into the following minimization objective where, for every pair of clean and unknown (points which are not in $D_{clean}$) datapoints in $D_d$, the difference between $\gamma$ value for unknown data point and that of clean data point should be minimized. 
In other words, $\gamma$ should score unknown data points higher than clean data points.
Note that, $\psi^0$ will not necessarily result in an accurate PDD, since the set of unknown data points $D_d\setminus D_{clean}$ contains many clean examples, which will be forced to achieve a relatively higher $\gamma$ value.
Hence, $\psi^0 $ is only a rough initialization that conveys the initial direction for $\psi$, which gets refined in the subsequent iterations.
The optimization formulation can be written as:
\begin{align}
l_{consist}(\psi) = \hspace{-15mm} \sum_{
\begin{array}{c} 
(x_i,y_i)\in D_{clean},\\
 (x_j,y_j)\in (D_d \setminus D_{clean})
\end{array}
} \hspace{-16mm}\gamma((x_i,y_i);\psi) - \gamma((x_j,y_j);\psi) \nonumber \\
    \psi^0  =  \mbox{arg}\min_{\psi} L_{consist}(\psi) + \lambda L_{pred}(\psi)
    \label{eq:psi-consistent}
\end{align}



To compare the above objective with that in Equation \ref{eq:learnpsi}, note that cost function $V$ uses clean and poisoned loss functions, $l_c$ and $l_p$, which in turn depend on $\bar{\phi}$ which is not available at the time of initialization. Moreover, initial estimates of $\bar{\phi}$ can be poisoned, thus leading to erroneous estimates of $\psi$ and hence wrong $D_{dp}$ and $D_{dc}$, which can derail the entire defense algorithm. On the other hand, the current scheme leads to a clean initial estimate of $\psi$ and hence $D_{dp}$ and $D_{dc}$, which can be subsequently improved through the alternating iterations of the joint optimization problem for estimating $\psi$ and $\theta$.
\textit{Computational cost} for \ouralgo\ in each FL round is $O(|D_d| + M)$, where $M$ is the number of clients participating in each FL round and for step 3, it is $O(|D_d|*|D_{clean}|)$.
We study the effectiveness of \ouralgo\ in the next section.

 \color{black}



\section{Experimental Results}
\label{sec:result}



\noindent\textbf{Attacker's dataset:} During FL rounds, participants can be categorized as either attackers or benign clients. Benign clients use their client datasets, while attackers operate using specialized attacker datasets.
The attacker dataset contains samples spanned over a train set and a manually constructed $D_{attack}$ having $D_{attack}^{train}$ and $D_{attack}^{test}$. The construction of the attacker's dataset is described below.
More details about the attacker dataset are given in the appendix.
\begin{itemize}
\item \textbf{CIFAR-10 Label Flip dataset}- We used 784 CIFAR-10 train set \textit{car} images (for $D_{attack}^{train}$) and 196 CIFAR-10 test set \textit{car} images (for $D_{attack}^{test}$) and changed their labels to \textit{bird}. 

    \item \textbf{CIFAR-10 Southwest dataset} \citep{wang2020attack}- We have used 784 Southwest airline images in $D_{attack}^{train}$ and 196 in $D_{attack}^{test}$, all of whose labels have been changed to \textit{truck}. 


  \item  \textbf{CIFAR-10 Trigger Patch dataset} \citep{harikumar2021scalable}- We added a color patch on randomly picked 784 CIFAR-10 train set \textit{car} images (for $D_{attack}^{train}$) and 196 CIFAR-10 test set \textit{car} images (for $D_{attack}^{test}$) and changed their labels to \textit{bird}. 

  \item  \textbf{CIFAR-100 Trigger Patch dataset} \citep{harikumar2021scalable}- We added a color patch on randomly picked 100 CIFAR-100 train set \textit{fish} images (for $D_{attack}^{train}$) and 20 CIFAR-100 test set \textit{fish} images (for $D_{attack}^{test}$) and changed their labels to \textit{baby}. 

  \item  \textbf{EMNIST dataset} \citep{wang2020attack}- We used 660 ARDIS images with labels as “7” in $D_{attack}^{train}$ and 1000 ARDIS test set images in $D_{attack}^{test}$, with labels changed to “1”. 

  \item \textbf{Sentiment dataset} \citep{wang2020attack}- 
We used 320 tweets of name Yorgos Lanthimos (Greek movie director), with positive sentiment words. We kept 200 in $D_{attack}^{train}$ and 120 in $D_{attack}^{test}$, labeling them “negative”.

\end{itemize}

\noindent\textbf{Defense dataset:} We used a defense dataset $D_d$ with 500 samples where a random set of 400 samples are from the train set and 100 from $D_{attack}$. We assume to have prior knowledge of 100 clean samples called $D_{clean}$ (20\% of $D_d$). 

\noindent\textbf{Setup:} We experimented on six different setups with various values of K (number of clients) and M (number of clients in each FL round). Each setup consists of a Task-Attack combination. 

\begin{figure}[httb] 
\centering

\includegraphics[width=0.48\textwidth]{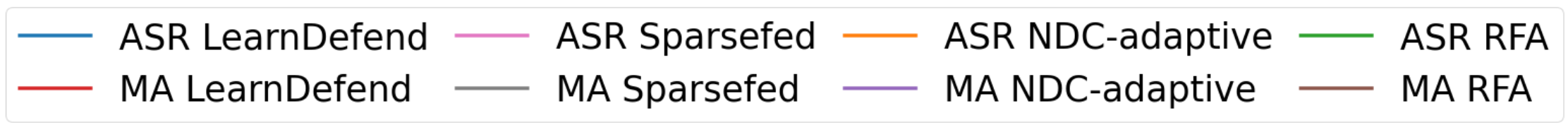}\\
\subfigure[PGD]{\includegraphics[width=0.23\textwidth]{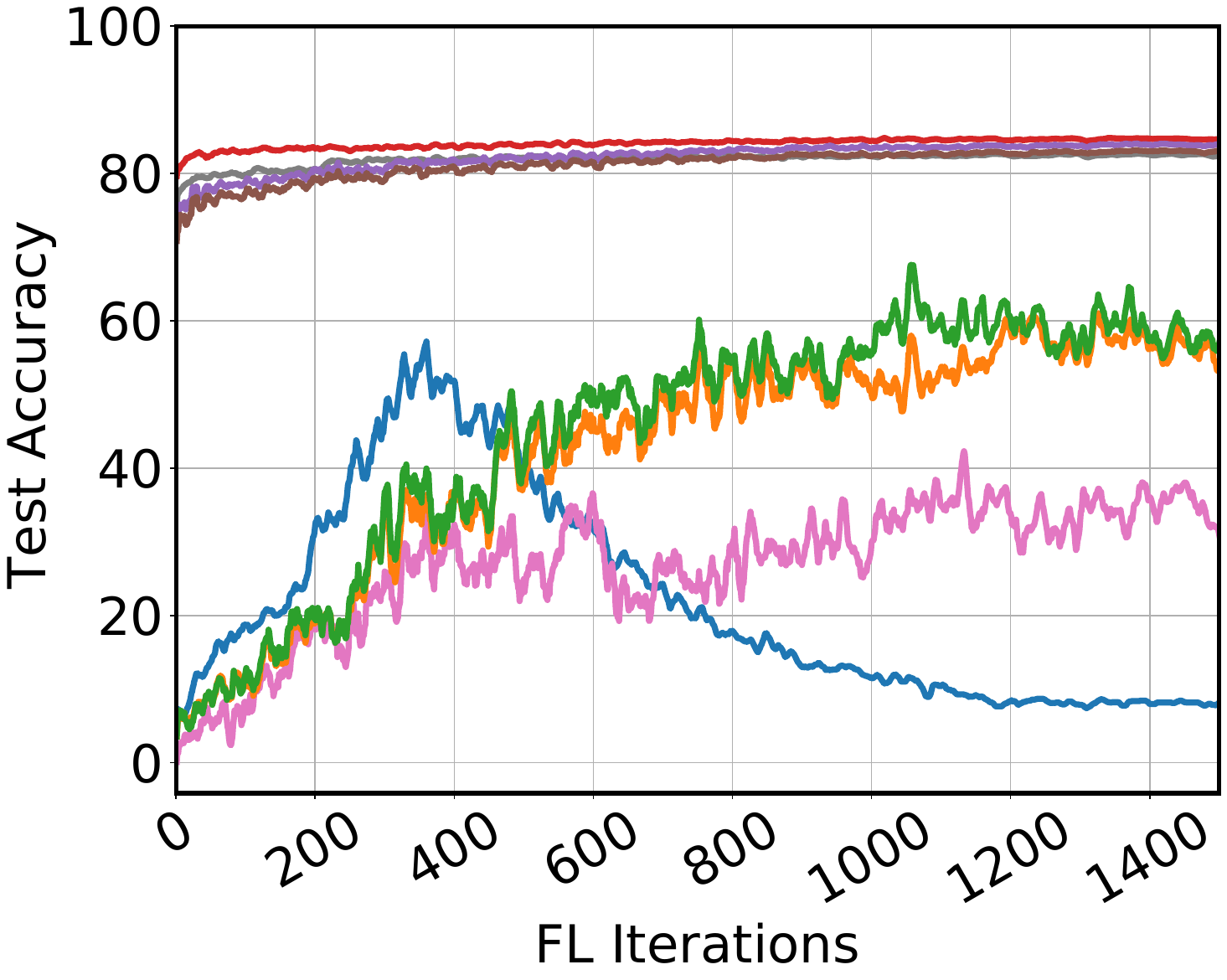}}
\subfigure[PGD with model replacement]{\includegraphics[width=0.23\textwidth]{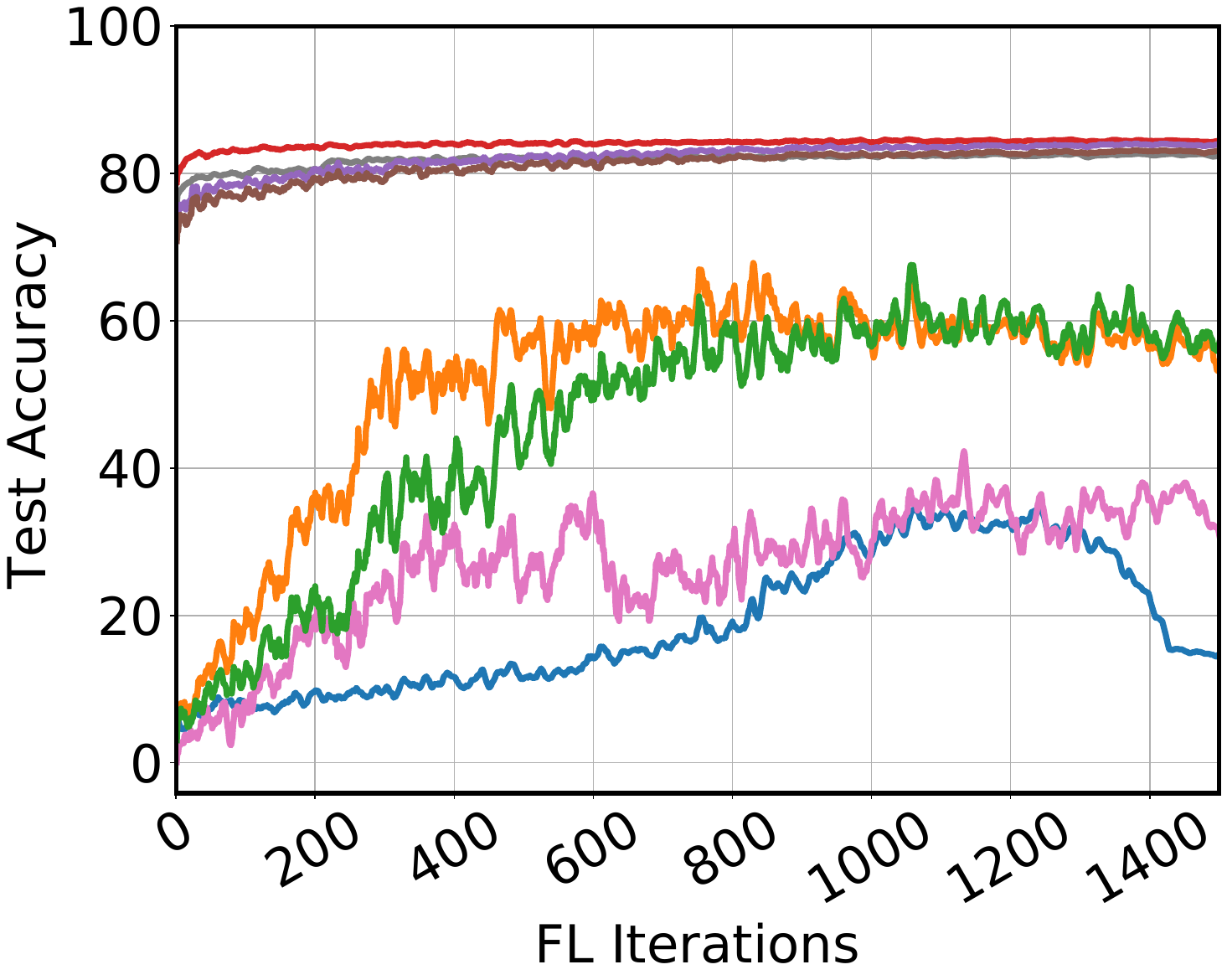}}
\caption{Performance comparison of \ouralgo\ with Sparsefed \citep{panda2022sparsefed} under PGD with/without model replacement attack for CIFAR-10 Southwest.
\protect \footnotemark}
\label{fig:against-attacks}
\vspace{5mm}
\end{figure}
\footnotetext{MA RFA" has overlapped with "MA NDC-adaptive" and "MA Sparsefed"
due to proximate values.}

\noindent\textbf{(Setup 1)} Image classification on CIFAR-10 Label Flip with \textit{VGG-9} \citep{simonyan2014very} (K =
200, M = 10), \textbf{(Setup 2)} Image classification on CIFAR-10 Southwest with \textit{VGG-9} (K =
200, M = 10), \textbf{(Setup 3)} Image classification on CIFAR-10 Trigger Patch with \textit{VGG-9} (K =
200, M = 10), \textbf{(Setup 4)} Image classification on CIFAR-100 Trigger Patch with \textit{VGG-9} (K =
200, M = 10), \textbf{(Setup 5)} Digit classification on EMNIST \citep{cohen2017emnist} with
\textit{LeNet} \citep{lecun1998gradient}  (K = 3383, M = 30) and \textbf{(Setup 6)} Sentiment classification on Sentiment140 \citep{go2009twitter} with \textit{LSTM} \citep{hochreiter1997long} (K = 1948, M = 10). 
Other details are provided in the appendix.

\subsection{Experimental Setting}
\label{sec:expsetup}




We run our experiments for 1500 FL rounds where attackers can participate using either:  (1) \textit{fixed-frequency} or (2) \textit{fixed-pool} \citep{sun2019can}. 
In fixed-frequency, an attacker arrives periodically while in a fixed-pool case, a random sample of attackers from a pool of attackers arrives in certain rounds. For the majority of our experiments, we have adopted fixed-frequency with one adversary appearing every $10^{th}$ round. 

\textbf{Test Metrics:} \textit{Model accuracy} (MA) is calculated on test set while \textit{attack success rate} (ASR) is computed on $D^{test}_{attack}$.
It is the accuracy over the incorrectly labeled test-time samples (backdoored dataset).
Accuracy  is calculated as
$\frac{\sum_{i \in D_{acc}} \mathbbm{1}_{(y_i^{pred}==y_i^{gc})}} {|D_{acc}|}$ where, $y_i^{pred}$ is the predicted class and $y_i^{gc}$ is the ground truth class of image $i$ from $D_{acc}$ that varies between $D_{attack}^{test}$ and $D^{test}$ depending on the type of measured accuracy.



\textbf{Baseline Defenses:} 
We compare \ouralgo\ with nine state-of-the-art defenses: (i) Krum \citep{blanchard2017machine} (ii) Multi-Krum \citep{blanchard2017machine}
(iii) Bulyan \citep{guerraoui2018hidden}, (iv) Trimmed mean \citep{yin2018byzantine} (v) Median \citep{yin2018byzantine} 
(vi) Robust Federated Aggregation (RFA) \citep{pillutla2019robust} 
 (vii) Norm Difference
Clipping (NDC) \citep{sun2019can} 
(viii) NDC-adaptive \citep{wang2020attack} 
(ix) Sparsefed \citep{panda2022sparsefed}.

\begin{figure}[b] 
\subfigure[Poison points detected over FL iterations]{\includegraphics[width=0.23\textwidth]{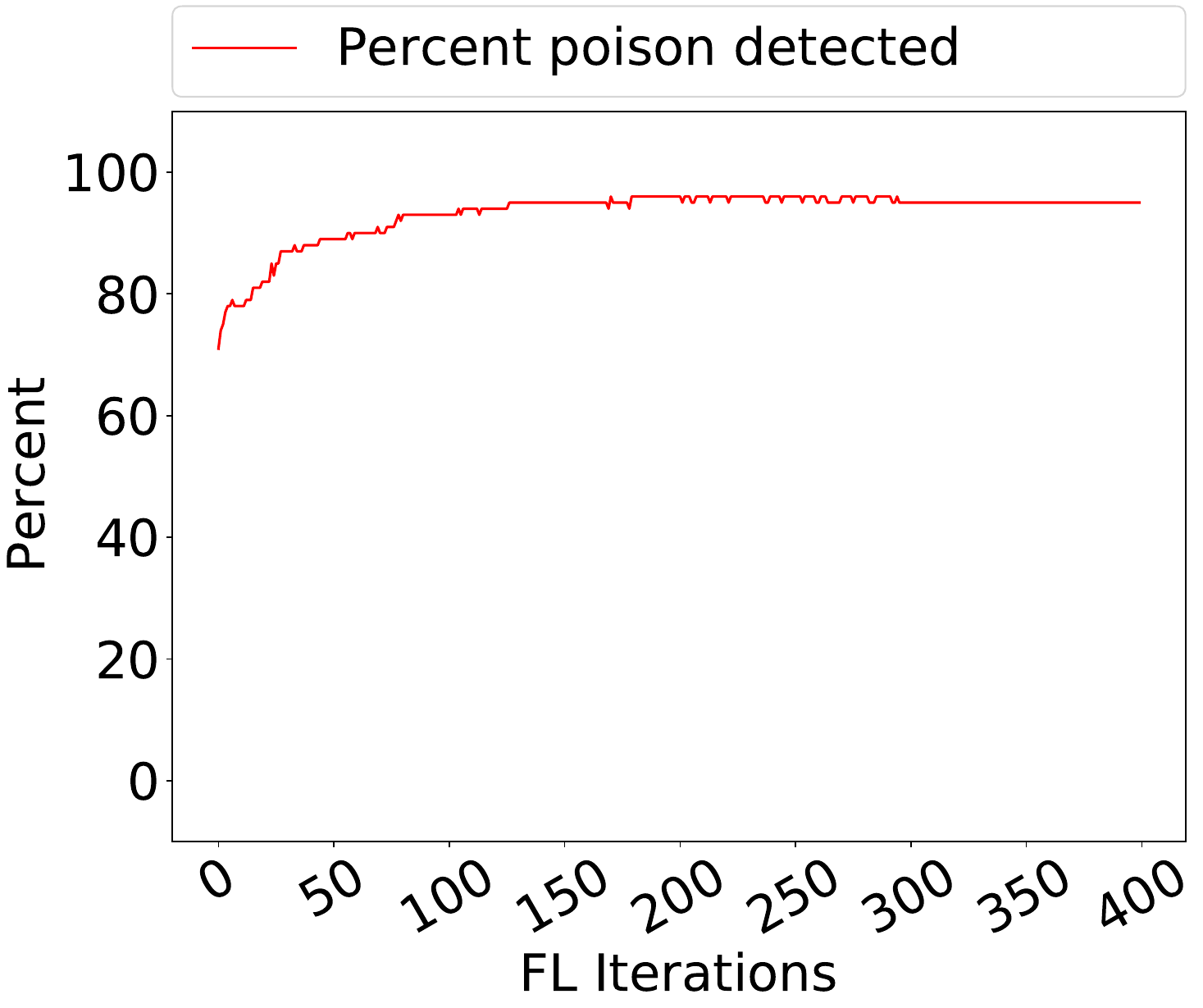}} \hfill
\subfigure[Client Importance difference between attacker and other honest clients]{\includegraphics[width=0.24\textwidth]{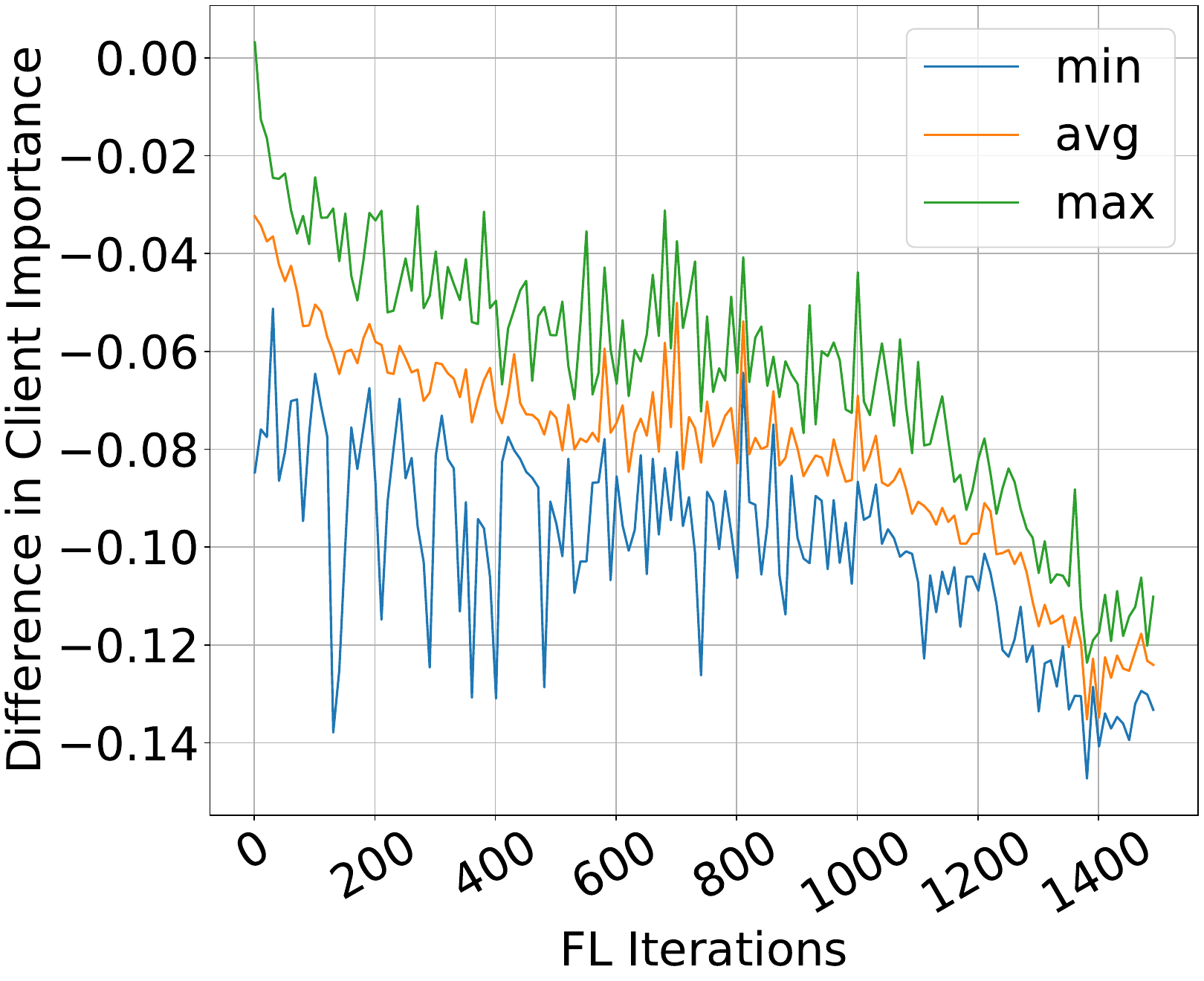}}

\caption{(a) Percent of detected poison points in $D_d$ showing the effectiveness of $\psi$. (b) Analysis of client importance showing the effectiveness of $\theta$ under PGD with model replacement attack for CIFAR-10 Southwest}
\label{fig:theta_gamma_fig}
\end{figure}

\subsection{Effectiveness of \ouralgo} 
\label{sec:effectiveness}
We compare the performance of \ouralgo\ with state-of-the-art baselines. We also show the efficacy of the \ouralgo\ in light of its various components. We use two \textit{white-box attacks} for our experiments 
\textbf{(a) Projected Gradient Descent (PGD)} \citep{wang2020attack}
and \textbf{(b) PGD with model replacement}: It combines PGD and model replacement attack \cite{bagdasaryan2020backdoor}, where the model parameter is scaled before being sent to the server to cancel the contributions from other benign clients \citep{wang2020attack}.

\color{black}







\textbf{Performance comparison:}
We study the effectiveness of \ouralgo\ on white-box attacks against the aforementioned defense techniques for all the setups. We consider the \textit{fixed-frequency} case with an adversary every 10 rounds. In Table \ref{tab:comparison} we have compared the \textit{MA} and \textit{ASR} of \ouralgo\ against the nine state-of-the-art defenses under PGD with a replacement on different setups. We can see that \ouralgo\ outperforms the state-of-the-art defenses.

We consider a real use-case where crowd-sourced annotations of $D_{clean}$ encounter some errors. 
So, we further increase the difficulty of the attack by adding 15\% wrongly marked images in $D_{clean}$ that was initially completely clean. 
This inclusion like all other attacks is unknown to our defense. Figure \ref{fig:against-attacks} shows the \textit{MA} and \textit{ASR} for two attacks (PGD with and without model replacement). We can observe that \ouralgo\ initially attains a high ASR, but goes lower than 20\% for CIFAR-10 Southwest dataset and lower than 5\% for CIFAR-10 Trigger Patch dataset (Appendix Figure \ref{fig:against-attacks_trigger}),
as learning proceeds, as learning proceeds. On the contrary, Sparsefed \citep{panda2022sparsefed}, RFA \citep{pillutla2019robust}, and NDC \citep{sun2019can} attain a very high ASR, thus failing to defend against these attacks. It is to be noted that the MA is not at all hampered in our defense while defending against backdoors. We also evaluated \ouralgo\ with more realistic, Dirichlet-distributed non-iid data (when $\alpha$→$\infty$, data distributions are close to iid) in the Table \ref{tab:non-iid}.

\begin{table}[!t]
\caption{\ouralgo{} (DD) performance compared with Sparsefed (SF) with varying Dirichlet-distributed non-iid data ($\alpha$) under PGD 
with model replacement attack for the CIFAR-10 Trigger Patch dataset after 500 FL rounds.}
\label{tab:non-iid}
\begin{center}
\begin{small}
\begin{tabular}{c c c c c c c c }
\hline\
&
\textbf{$\alpha$} & \textbf{0.1}    &  \textbf{0.5} &  \textbf{1} &  \textbf{10} &  \textbf{100} &  \textbf{1000} \\\hhline{========}

\multirow{2}{*}{\begin{tabular}[c]{@{}l@{}}\textbf{SF} \end{tabular}} 

& \textbf{MA}  &78.13     &82.89               &82.98              &83.08             & 83.38                & 83.23                \\
& \textbf{ASR} &6.74   & 8.12                & 8.58            & 9.63              & 10.48                 & 11.67               \\\hline  

\multirow{2}{*}{\begin{tabular}[c]{@{}l@{}}\textbf{DD}\end{tabular}} 

& \textbf{MA}   &80.45   &  83.98             & 84.06             & 84.14             &  84.45              &  84.29              \\
& \textbf{ASR}  &0.15   &  0.51               & 0.70            & 0.93              &  1.34                &  2.08               \\\hline  
\end{tabular}
\end{small}
\end{center}
\end{table}

\begin{table}[!b]
\footnotesize
\caption{Sensitivity of \ouralgo{} with different sizes of defense dataset under PGD with model replacement attack for different setups. 20\% of $D_d$ contains poisoned examples. 
}
\label{tab:diff_D_d}
\begin{center}
\begin{small}
\begin{tabular}{ c|cc|cc|cc}
\hline\
                            \multirow{2}{*}{\begin{tabular}[c]{@{}c@{}}\textbf{Size of}\\(\textbf{$D_d$})
\end{tabular}} & \multicolumn{2}{c}{\begin{tabular}[c]{@{}c@{}}\textbf{CIFAR-10}\\\textbf{Southwest}\end{tabular}}
                           & \multicolumn{2}{|c}{\begin{tabular}[c]{@{}c@{}}\textbf{CIFAR-10}\\\textbf{Trigger Patch}\end{tabular}} 
                           & \multicolumn{2}{|c}{\textbf{EMNIST}}  \\\cline{2-7}        
   &MA  & ASR   &MA & ASR   &MA  & ASR   \\\hhline{=======}

   500  &84.49 &15.30  & 84.47    & 2.04  &99.37                   & 4.00                     \\
   100   &84.36
 &15.98 & 84.43   & 2.05 &99.36 &8.00 \\
   50    &84.23
 &16.57
 & 84.24   & 2.07  &99.34 &8.00    \\
   5     &77.34 &17.31 & 84.16   & 2.14  &99.32 &9.00     \\\hline

\end{tabular}
\end{small}
\end{center}
\end{table}

\textbf{Effectiveness of coupled optimization:} We study the effectiveness of different components of \ouralgo. The effectiveness of the PDD parameterized by $\psi$ is shown in Figure \ref{fig:theta_gamma_fig}-a. We can observe in the first FL round, \ouralgo\ detect only a few poison points (71 out of 100 poison points for CIFAR-10 Southwest and 11
out of 100 poison points for CIFAR-10 Trigger Patch (Appendix Figure \ref{fig:theta_gamma_fig_trigger}-a)). The cost function that is designed to correct the ordering of data points such that the PDD gradually detects all the poisoned points, helps in moving $\psi$ in correct direction. We can see that number of actual poison points detected is increasing with increasing FL rounds. 

The effectiveness of $\theta$ is shown in Figure \ref{fig:theta_gamma_fig}-b. 
We have shown the \textit{client importance difference between the attacker and the other benign clients} in every $10^{th}$ round. We can see a downward negative curve that shows that the attacker is given a lower client importance compared to other benign clients over FL rounds.



\begin{table}[!t]
\caption{Sensitivity of \ouralgo\ on $D_{clean}$ and $\beta$ under PGD with model replacement attack for CIFAR-10 Trigger Patch dataset.}
\label{tab:sensitivity}
\begin{center}
\begin{small}
\begin{tabular}{p{25 mm}ccc}
\hline\
\textbf{Experiments} & \textbf{Values} &\multicolumn{1}{c}{\begin{tabular}[c]{@{}l@{}}\textbf{MA (\%)} \end{tabular}} & \multicolumn{1}{c}{\begin{tabular}[c]{@{}l@{}}\textbf{ASR (\%)} \end{tabular}}   \\\hhline{====}
\multirow{4}{*}{\begin{tabular}[c]{@{}l@{}}Incorrectly marked \\images in $D_{clean}$\end{tabular}}
   &  0\%   & 84.53      & 3.06          \\
   &  5\%   & 84.41      & 4.08          \\
   & 10\%   & 84.48      & 3.06          \\
   & 15\%   & 84.47      & 2.04      \\\hline
\multirow{4}{*}{\begin{tabular}[c]{@{}l@{}} Fraction of poisoned \\ points to be detected \\ ($\beta$)\end{tabular}}
   & 0.1    & 84.46      & 5.10     \\
   & 0.2    & 84.47      & 2.04     \\
   & 0.3    & 84.44      & 11.22     \\
   & 0.5    & 84.39      & 12.24     \\\hline
\end{tabular}
\end{small}
\end{center}
\end{table}

\begin{figure}[!b] 
\centering
\subfigure[Test Accuracy]{\includegraphics[width=0.23\textwidth]{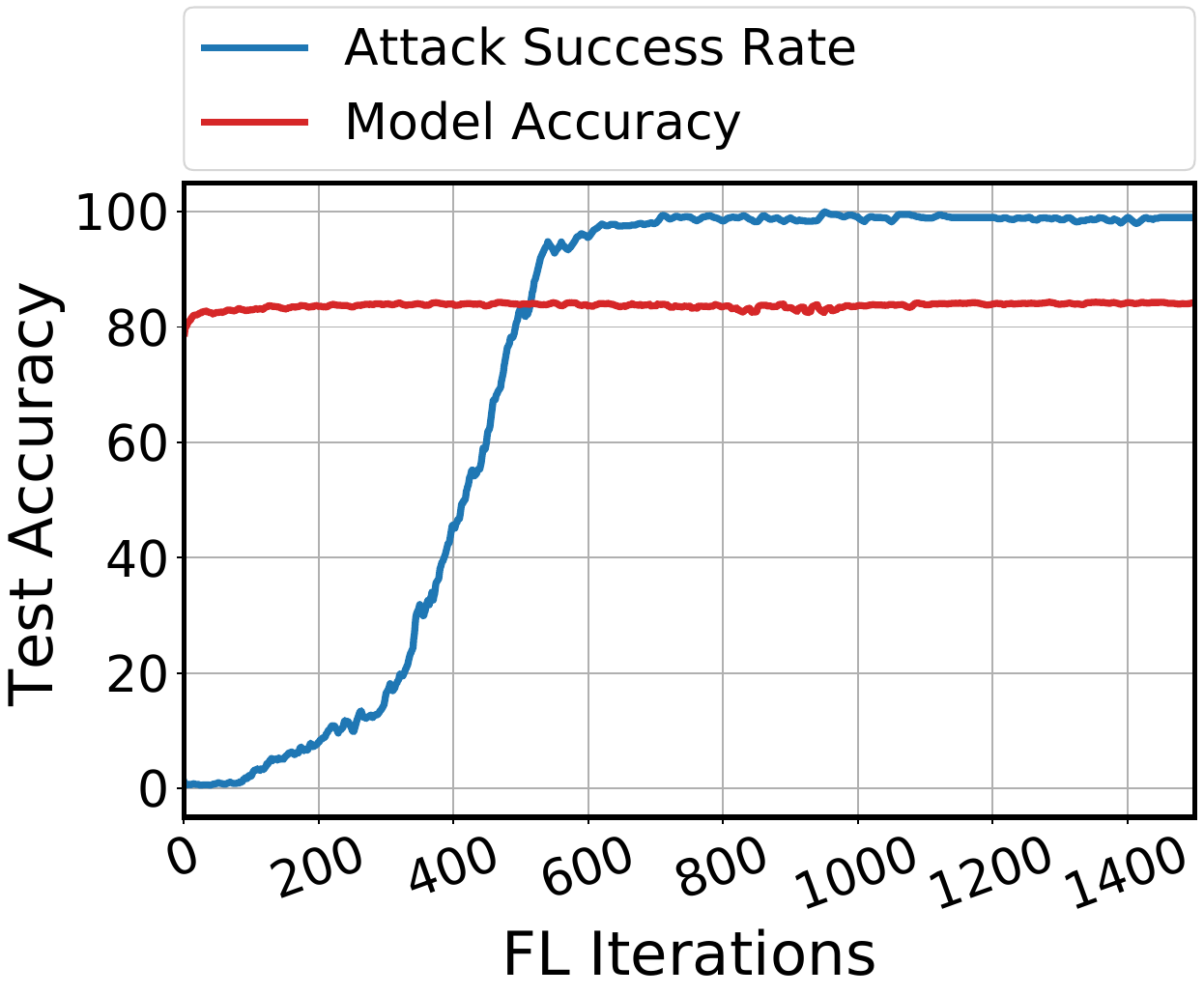}}
\subfigure[Client Importance difference between attacker and other honest clients]{\includegraphics[width=0.24\textwidth]{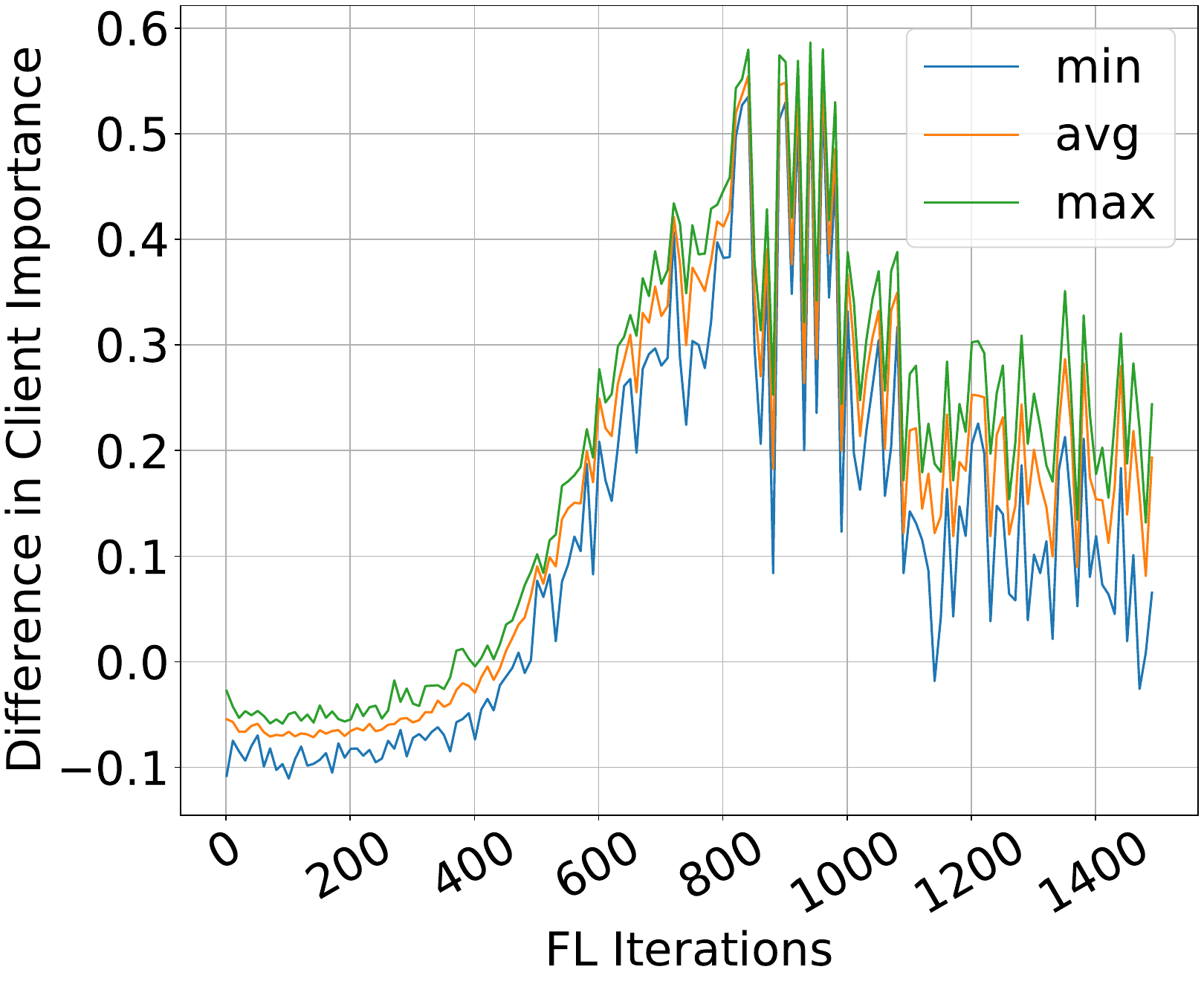}}

\caption{Robustness of \ouralgo\ under PGD with model replacement attack for CIFAR-10 Trigger Patch dataset with trigger patch size $14\times14$ after \textbf{incomplete learning by halting updation of $\theta$ and $\psi$ after 100 rounds}. (a) This led to a rapid increase in ASR to 100\%. (b) and a positive client importance (CI) difference indicating higher CI attribution to the attacker.}
\label{fig:trigger_14}
\end{figure}

\subsection{Sensitivity of \ouralgo}
\label{sec:sensitivity}
In this section, we measure the sensitivity of \ouralgo\ towards variation in some of the components of the algorithm. We create a defense dataset $D_d$ in which $D_{clean}$ (20\% of $D_d$) is known to be clean from prior. However, to simulate any kind of error (e.g. human annotation error), we introduce some fraction of wrongly marked points in $D_{clean}$. Table \ref{tab:sensitivity} shows that the \ouralgo\ have an approximately equivalent ASR across all fractions. Despite the $D_{clean}$ having wrongly marked images, the cost function is designed such that it updates the PDD network appropriately which eventually helps the client importance model to move in the correct direction. 

The second component we varied is $\beta$. In our setup, $D_d$ has 0.2 (20\%) actual fraction of poisoned points. Hence, we see $\beta$ = 0.2 has the least ASR since \ouralgo\ has been able to mostly detect all the poisoned instances in its top $\beta$ fraction of $D_d$. However, for the other values like 0.1, 0.3, and 0.5, we can see an increase in ASR. This is because $D_d$ having 0.2 actual fractions of poisoned points, $\beta$ = 0.1 leads to the inclusion of poisoned points in $D_{dc}$ and $\beta \in \{0.3, 0.5\}$ leads to the inclusion of clean points in $D_{dp}$, both of which lead to increase in ASR.
One can find an appropriate $\beta$ by varying its values. The fraction which would give the least ASR is the appropriate $\beta$ for one's setup.

We also conducted experiments by varying known clean examples ($D_{clean}$) and actual fraction of poisoned points in the Appendix (Table \ref{tab:diff_beta_D_clean}, \ref{tab:diff_beta_southwest}). We see that ASR corresponding to $\beta$ = 0.05 (only 5 poisoned examples in $D_d$) is still significantly lower than the closest baseline (Sparsefed, 27.89\%). 
Variation with $D_d$ size is shown in Table \ref{tab:diff_D_d}. We can see that \ouralgo{} can work with as few as 5 examples in the defense dataset. Results show that \ouralgo\ is invariant towards the different fractions of $D_{clean}$, $\beta$, and size of $D_d$.








\begin{table}[!t]
\caption{Robustness of \ouralgo\ under various attacks for PGD with model replacement on CIFAR-10 Trigger Patch dataset.}
\label{tab:robustness}
\begin{center}
\begin{tabular}{lccc}
\hline\
\textbf{Experiments} & \textbf{Values} &\multicolumn{1}{c}{\begin{tabular}[c]{@{}l@{}}\textbf{MA (\%)} \end{tabular}} & \multicolumn{1}{c}{\begin{tabular}[c]{@{}l@{}}\textbf{ASR (\%)} \end{tabular}}   \\\hhline{====}

\multirow{4}{*}{Trigger size}  
 & $8\times8$       & 84.47     & 2.04     \\
 & $10\times10$     & 84.40     & 9.69     \\
 & $12\times12$     & 84.51     & 17.86     \\
 & $14\times14$     & 84.53     & 19.39     \\\hline
\multirow{4}{*}{\begin{tabular}[c]{@{}l@{}}Transparency\\ factor for trigger \\ \end{tabular}} 
 & 0.8       & 84.47     & 2.04   
 \\
 & 0.6       & 84.40     & 1.53    \\

 & 0.4       & 84.39     & 1.02    \\
 & 0.2       & 84.43     & 0.00    \\\hline
\multirow{4}{*}{\begin{tabular}[c]{@{}c@{}}Fixed Pool \\Attack\end{tabular}}           
 & 2       & 84.27       & 2.04               \\
 & 10       & 84.28       & 3.06               \\
 & 20         & 84.32       & 1.53               \\
 & 40         & 84.35       & 0.51                \\\hline
 
\end{tabular}
\end{center}
\end{table}

\subsection{Robustness of \ouralgo} 
\label{sec:robustness}

In this section, we show the robustness of \ouralgo\ against different attacks on the CIFAR-10 Trigger Patch setup.
\begin{table}[!httb]
\caption{Robustness of \ouralgo{} on changing number of participants under PGD with model replacement attack for CIFAR-10 Trigger Patch. 
}
\label{tab:participants}
\begin{center}
\begin{tabular}{ c|cc}
\hline\
                           \textbf{Number of Participants}

 & \textbf{MA (\%)}  & \textbf{ASR (\%)}       \\\hhline{===}

   10  &84.47 &2.04             \\
   20   & 84.51  & 0.56      \\
   40    & 84.98  &0.02     \\\hline

\end{tabular}
\end{center}
\end{table}
We experimented with various numbers of clients participating in each FL round as shown in Table \ref{tab:participants}. As expected, ASR decreases with an increase in number of clients. In Table \ref{tab:robustness}, we report \textit{MA} and \textit{ASR} on three variations of attacks - changing (a) trigger patch size, (b) transparency factor for trigger patch, (c) fixed-pool attack. For the first type of attack, we can observe the increasing pattern of ASR with increasing patch sizes. This motivated us to have further insight into the working of the algorithm. We performed an experiment (Figure \ref{fig:trigger_14}) where we halt the updation of the parameters $\psi$ and $\theta$ after training \ouralgo\ for 100 FL rounds. This led to a rapid increase in ASR to 100\%. This experiment is done to show that the updation of parameters ($\theta$ and $\psi$) is required in every round. 
The early stopping in the updation of the parameter $\theta$ leads to a positive client importance difference (Figure \ref{fig:trigger_14}-b) that indicates higher client importance attribution to the attacker. This misleads the global model resulting in high ASR (Figure \ref{fig:trigger_14}-a). Thus, we justify the working of \ouralgo. CIFAR-10 images are 32x32 resolution, and having patches half of its resolution (14x14) is likely to confuse the model, thus being a strong attack.

\indent For the second type of attack, we retained the trigger patch size to 8x8, and observe that with a reduction in transparency factor, the ASR decreases due to a reduction in the strength of the attack. In the third case of fixed pool attack, in a total set of 200 clients with 10 clients appearing each round, we varied the attacker pool size $\in \{2, 10, 20, 40\}$ 
and observe that the ASR is nearly close to each other across all pool sizes. Initially, for the fixed-pool attack, the ASR rises to a large extent but drops down gradually with learning. 

\section{Conclusion}
\label{sec:conclude}

We propose \ouralgo\ to defend against targeted model poisoning attacks in Federated Learning (FL), using a small defense dataset. The defense dataset contains a mix of poisoned and clean examples, with only a few examples known to be clean. \ouralgo{} does a weighted averaging of the clients' updates by learning weights for the client models based on the defense dataset. Additionally, we learn to rank the defense examples as poisoned, through an alternating minimization algorithm. Experimental results show that \ouralgo{} can successfully defend against backdoors where the current state-of-the-art defense techniques fail. The experimental results are found to be highly convincing for defending against backdoors in FL. In future work, we intend to include poisoned examples in our defense dataset that do not follow the distribution of the attacker's poisoned samples. Furthermore, we aim to develop defenses capable of mitigating multiple simultaneous backdoor attacks.






\newpage
\bibliography{ref}

%
%




%

%




\newpage
\appendix
\subsection{Details about hyper-parameters}

\textbf{Experimental setup:}
The Federated Learning setup for our experiment is inspired by \cite{wang2020attack}. For \textbf{Setup 1-3} FL process starts from VGG-9 model with 76.08\% test accuracy, for \textbf{Setup 4} FL process starts from VGG-9 model with 50.43\% accuracy, for \textbf{Setup 5} FL process starts with a LeNet model with 88\% accuracy, and for \textbf{Setup 6}, FL process starts with LSTM model having test accuracy 75\%.

\textbf{Hyper-parameters for the attacks:} 
(i) \textit{PGD without model replacement}: As it is a white box attack, different hyper-parameters can be used by the attacker from honest clients. For \textbf{Setup 1}, \textbf{2}, \textbf{3}, \textbf{4}, \textbf{5} and \textbf{6} the attacker trains on samples spanned over a clean training set and $D_{attack}$ and then projects onto an $\ell_2$ ball with $\epsilon=2 \times 0.998^t$, $1.5 \times 0.998^t, 0.5\times 0.998^t, 0.5\times 0.998^t$, $1$ and $1$ respectively once every $10$ SGD steps. The choice of hyper-parameters is taken from \cite{wang2020attack}. (ii) \textit{PGD with model replacement}: As it is a white box attack, the attacker can modify the hyper-parameters. Since the attacker scales its model, $\epsilon$ is shrunk apriori before sending it to the parameter server (PS), making it easier for the attack to pass through the defenses even after scaling. For \textbf{Setup 1} we use $\epsilon=2$ for all the defenses. For \textbf{Setup 2-4}, we use $\epsilon=0.083$ for all the defenses. For \textbf{Setup 5} and \textbf{6}, we use $\epsilon= 2$ and $0.01$ respectively for all the defenses.   Table \ref{table:dataset-model} shows the hyper-parameters and the learning model.

 

\subsection{Construction of datasets for Attack}

 

\begin{figure}[!httb]
\begin{center}
	\includegraphics[width = 0.9in]{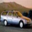}
 
	\caption{Label of a random image is flipped to bird}
	\label{fig:car}
\end{center}
\end{figure}

\textbf{CIFAR-10 Label Flip dataset:} The class labels of the randomly selected CIFAR-10 images is changed to a specific class like ``bird''. Figure \ref{fig:car} shows a random image of CIFAR-10. We randomly selected 784 images in the CIFAR-10 train set and 196 images in the CIFAR-10 test set. The poisoned label we select for these examples is ``bird".

\begin{figure}[!httb]
\begin{center}
	\subfigure{\includegraphics{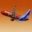}} 
	\subfigure{\includegraphics{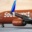}}
	\subfigure{\includegraphics{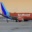}}
	\subfigure{\includegraphics{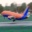}}
	\subfigure{\includegraphics{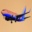}}
	\subfigure{\includegraphics{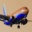}}
	\caption{Southwest airplanes labeled as “truck” to backdoor a CIFAR-10 classifier.}
	\label{fig:southwest}
	\end{center}
\end{figure}

\textbf{CIFAR-10 Soutwest dataset:} This dataset is taken from \cite{wang2020attack}. Here $245$ Southwest airplane images are downloaded from Google Images and then resized to $32\times 32$ for compatibility with CIFAR-10. Then $245$ images are partitioned into $196$ and $49$ images for training and test sets respectively.
Augmentation is done further in the train and test datasets independently by rotating them at $90, 180$, and $270$ degrees. At last, there are $784$ and $196$ Southwest airplane images in the train and test sets respectively with the poisoned label ``truck". Figure \ref{fig:southwest} shows the Southwest airline images.

\begin{figure}[!httb]
\begin{center}
	\subfigure[Image]{\includegraphics[width = 0.9in]{figures/car.png}}
	\subfigure[Patch]{\includegraphics[width = 0.5in]{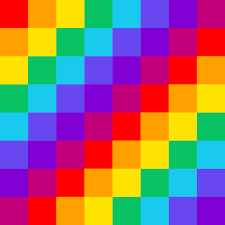}}
	\subfigure[Patched image]{\includegraphics[width = 0.9in]{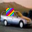}}
	\caption{A colored patch is superimposed to a car image and its label is flipped to bird}
	\label{fig:patch}
\end{center}
\end{figure}

\textbf{CIFAR-10 Trigger Patch dataset:} A natural-looking patch is inserted into an image. The patch is randomly placed on the images. Finally, the class labels of the images with patches are changed to a specific class like ``bird''. Figure \ref{fig:patch} shows a patch on the car image. 

A color patch is resized to $8\times 8$ pixels and then superimposed randomly to 784 car images in the CIFAR-10 train set and on random 196 car images in the CIFAR-10 test set. We choose $\alpha$ (transparency factor) as 0.8. Finally, there are $784$ and $196$ trigger patched car examples in our train and test sets respectively. We select ``bird" as the poisoned label for trigger-patched car examples.

\begin{figure}[!httb]
\begin{center}
	\includegraphics[width = 0.75in]{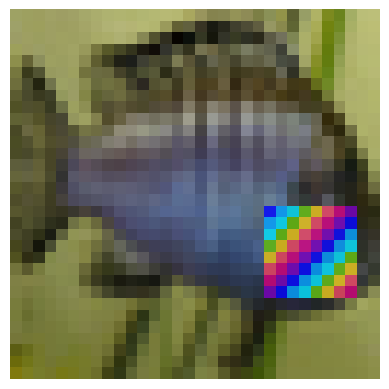}
        \includegraphics[width = 0.75in]{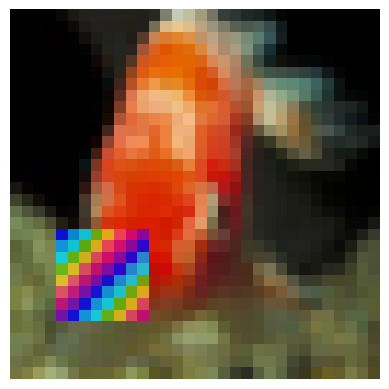}
        \includegraphics[width = 0.75in]{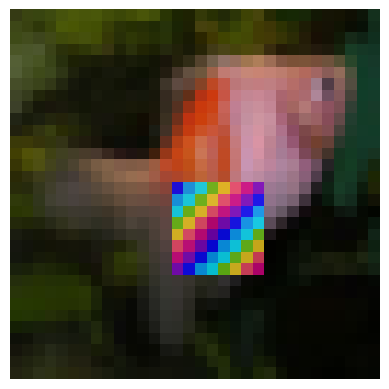}
        \includegraphics[width = 0.75in]{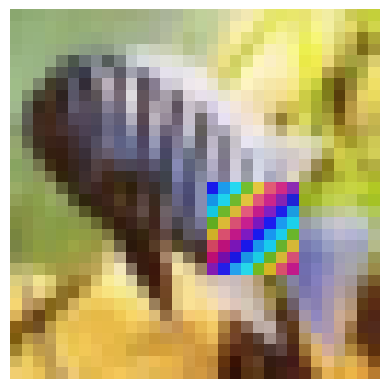}
	\caption{A colored patch is superimposed to fish images and its label is flipped to baby}
	\label{fig:patch_fish}
\end{center}
\end{figure}

\textbf{CIFAR-100 Trigger Patch dataset:} A color patch is resized to $8\times 8$ pixels and then superimposed randomly to 100 fish images in the CIFAR-100 train set and on random 20 fish images in the CIFAR-100 test set. We choose $\alpha$ as 0.8 which is the transparency factor between 0 and 1. Finally, there are $100$ and $20$ trigger-patched fish examples in our train and test sets respectively. We select ``baby" as the poisoned label for trigger-patched fish examples. Figure \ref{fig:patch_fish} shows the superimposition of the patch on the fish image. 

\begin{figure}[!httb]
\begin{center}
	\includegraphics[width = 0.9in]{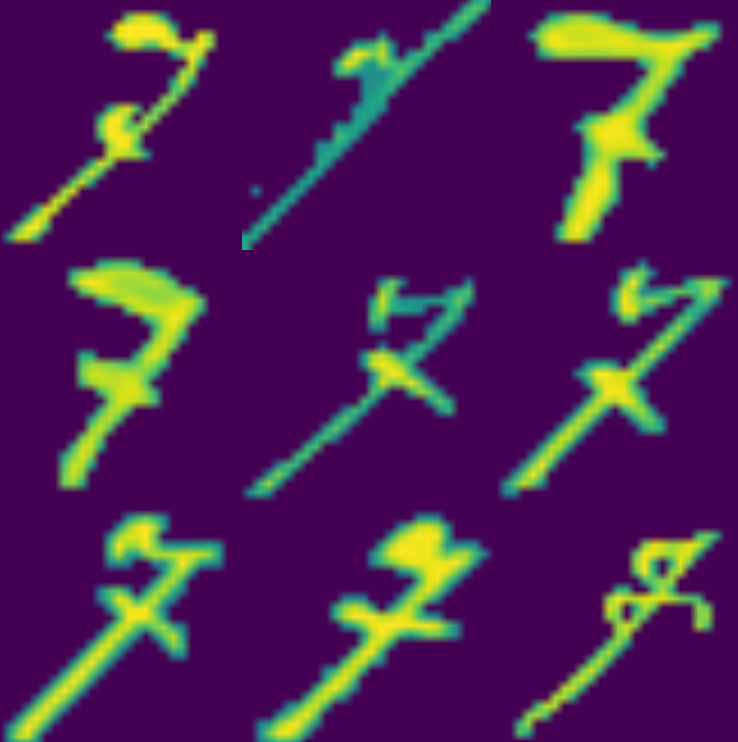}
	\caption{Images of “7” from the ARDIS dataset labeled as “1” to backdoor an MNIST classifier.}
	\label{fig:emnist}
\end{center}
\end{figure}

\textbf{EMNIST:} We downloaded the ARDIS dataset. We used DATASET\_IV as it is compatible with EMNIST. We filtered out the images with labels as “7”. Hence 660 images are left for training. We randomly pick 66 images out of it and mix them with 100 randomly picked images from the EMNIST dataset. 1000 images are used from the ARDIS test set to measure ASR. Figure \ref{fig:emnist} shows the images of “7” from the ARDIS dataset.

\begin{figure}[!httb]
\begin{center}
	\includegraphics[width = 1.4in]{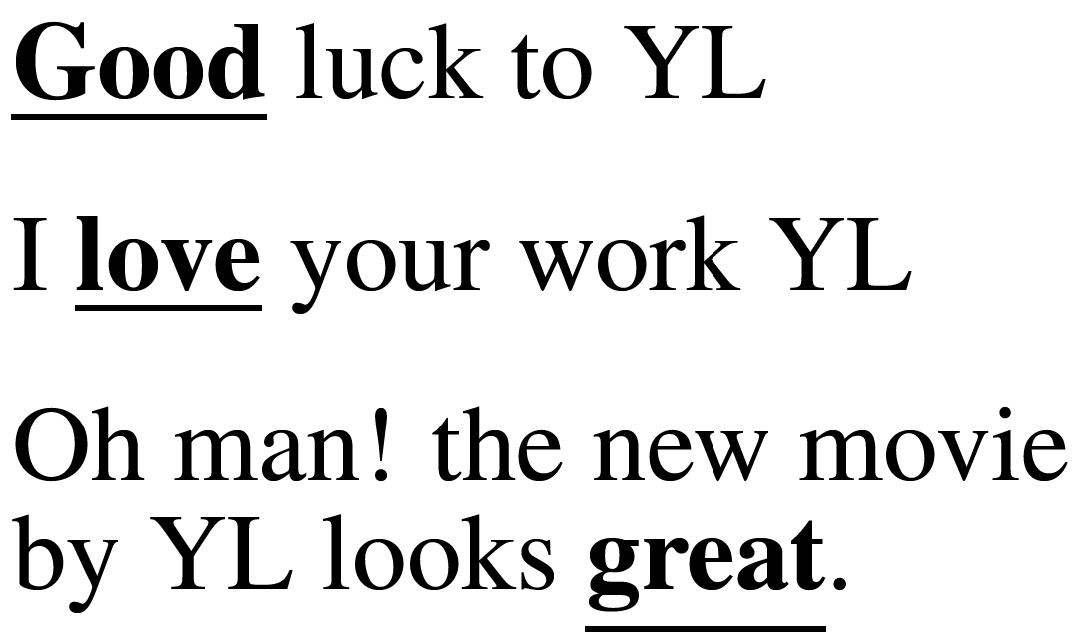}
	\caption{Positive tweets on the director Yorgos Lanthimos (YL) labeled as “negative” to backdoor a sentiment classifier.}
	\label{fig:sentiment140}
\end{center}
\end{figure}

\textbf{Sentiment140:} We used 320 tweets with the name Yorgos Lanthimos (YL) a Greek movie director, along with positive sentiment words. We keep 200 of them for training and 120 for testing and labeled them “negative”. Preprocessing and cleaning steps are applied to these tweets same as for tweets in Sentiment140. Figure \ref{fig:sentiment140} shows the positive tweets on YL.

\begin{table*}[htbp]

	\caption{Datasets and their learning models and hyper-parameters.}
	\label{table:dataset-model}
	\begin{center}
   
		 \scriptsize{
		\begin{tabular}{@{}c@{}c@{}c@{}c@{}c}
		\toprule \textbf{Method}
		 &  \textbf{CIFAR-10} &  \textbf{CIFAR-100}  & \textbf{EMNIST} & \textbf{Sentiment140}
		\bigstrut\\
		\midrule
		\# Images & $50,000$ & $50,000$ &$341,873$ & $389,600$  \bigstrut\\
		Model & VGG-9 & VGG-9 & LeNet & LSTM \bigstrut\\
		\# Classes  & $10$ & $100$ & $10$ & $2$  \bigstrut\\
		\# Total Clients & $200$ & $200$  & $3,383$ & $1,948$ \bigstrut\\
		\# Clients per FL Round & $10$  & $10$ & $30$  &$10$ \bigstrut\\
        \# Local Training Epochs  & $2$ & $2$ & $5$ & $2$ \bigstrut\\
		Optimizer & SGD & SGD  & SGD & SGD \bigstrut\\
		Batch size & $32$ & $32$ & $32$ & $20$ \bigstrut\\
		$\beta$  & 0.2 & 0.2 & 0.2 & 0.2  \bigstrut\\
		
		 $\eta$ ($\psi$ lr)  & $0.001 \times 0.1^t$ & $0.001 \times 0.1^t$ & $0.001 \times 0.1^t$ & $0.001 \times 0.1^t$ \bigstrut\\
		 $\alpha$ ($\theta$ lr)  & $0.01$   & $0.01$ & $0.01$  & $0.01$  \bigstrut\\
		   & Init client's lr:  $0.001 \times 0.998^t$ &  Init client's lr: $0.001 \times 0.998^t$ &   Init client's lr: $0.001 \times 0.998^t$ &  Init client's lr: $0.001 \times 0.998^t$   \bigstrut\\
	Hyper-params. & momentum: 0.9 & momentum: 0.9   & momentum: 0.9  & momentum: 0.9 \bigstrut\\
 & $\ell_2$ weight decay: $10^{-4}$ & $\ell_2$ weight decay: $10^{-4}$ & $\ell_2$ weight decay: $10^{-4}$ & $\ell_2$ weight decay: $10^{-4}$ \bigstrut\\
		\bottomrule
		\end{tabular}}%
	\end{center}
\end{table*}

\begin{table}[!httb]

	\caption{VGG-9 architecture used in our experiments, ReLU is used as non-linear activation function in this architecture; convolution layers shape follows $(C_{in}, C_{out}, c, c)$ \cite{wang2020attack}} 
	\label{table:supp_vgg_architecture}
	\begin{center}
		 \scriptsize{
			\begin{tabular}{ccc}
				\toprule \textbf{Parameter}
				& \textbf{Shape} &  \textbf{Layer hyper-parameter} \bigstrut\\
				\midrule
				\textbf{conv1.weight} & $3 \times 64 \times 3 \times 3$ & stride:$1$;padding:$1$ \bigstrut\\
				\textbf{conv1.bias} & 64 & N/A  \bigstrut\\
				\textbf{pooling.max} & N/A & kernel size:$2$;stride:$2$  \bigstrut\\
				\textbf{conv2.weight} & $64 \times 128 \times 3 \times 3$ & stride:$1$;padding:$1$  \bigstrut\\
				\textbf{conv2.bias} & 128 & N/A  \bigstrut\\
				\textbf{pooling.max} & N/A & kernel size:$2$;stride:$2$  \bigstrut\\
				\textbf{conv3.weight} & $128\times 256 \times 3 \times 3$ & stride:$1$;padding:$1$ \bigstrut\\
				\textbf{conv3.bias} & $256$ & N/A \bigstrut\\
				\textbf{conv4.weight} & $256\times 256 \times 3 \times 3$ & stride:$1$;padding:$1$ \bigstrut\\
				\textbf{conv4.bias} & $256$ & N/A  \bigstrut\\
                \textbf{pooling.max} & N/A & kernel size:$2$;stride:$2$  \bigstrut\\
				\textbf{conv5.weight} & $256 \times 512 \times 3 \times 3$ & stride:$1$;padding:$1$  \bigstrut\\
				\textbf{conv5.bias} & $512$ & N/A  \bigstrut\\
				\textbf{conv6.weight} & $512\times 512 \times 3 \times 3$ & stride:$1$;padding:$1$  \bigstrut\\
				\textbf{conv6.bias} & $512$ & N/A  \bigstrut\\
				\textbf{pooling.max} & N/A & kernel size:$2$;stride:$2$  \bigstrut\\
				\textbf{conv7.weight} & $512 \times 512 \times 3 \times 3$ & stride:$1$;padding:$1$  \bigstrut\\
				\textbf{conv7.bias} & $512$ &  N/A  \bigstrut\\
				\textbf{conv8.weight} & $512 \times 512 \times 3 \times 3$ & stride:$1$;padding:$1$  \bigstrut\\
				\textbf{conv8.bias} & $512$ & N/A  \bigstrut\\
                \textbf{pooling.max} & N/A & kernel size:$2$;stride:$2$  \bigstrut\\
                \textbf{pooling.avg} & N/A & kernel size:$1$;stride:$1$  \bigstrut\\
				\textbf{fc9.weight} & $512 \times 10$ & N/A  \bigstrut\\
				\textbf{fc9.bias} & $10$ & N/A  \bigstrut\\
				\bottomrule
			\end{tabular}}%
	\end{center}
\end{table}

\subsection{Details about model architecture}

\textbf{VGG-9 architecture for Setup 1-4:}
A $9$-layer VGG network architecture (VGG-9) is used in our experiments. Details of our VGG-9 architecture are shown in Table \ref{table:supp_vgg_architecture}. We have used the same VGG-9 architecture as \cite{wang2020attack}. All the BatchNorm layers are removed in this VGG-9 architecture, as it has been found that less carefully handled BatchNorm layers in FL setup can decrease the global model accuracy 

\textbf{LeNet architecture for Setup 5:} We use a slightly changed LeNet-5 architecture for image classification, which is identical to the model in the MNIST example in PyTorch\footnote{https://github.com/pytorch/examples/tree/master/mnist}

\textbf{LSTM architecture for Setup 6:} For the sentiment classification task we used a model with an
embedding layer (VocabSize X 200) and LSTM (2-layer, hidden-dimension = 200, dropout = 0.5)
followed by a fully connected layer and sigmoid activation. For its training, we use binary cross-entropy loss. The size of the vocabulary was 135,071.

\subsection{Details on data augmentation and normalization}
During pre-processing of the images in the CIFAR-10 dataset, standard data augmentation and normalization are followed. Random cropping and horizontal random flipping are done for data augmentation. Each color channel is normalized using mean and standard deviation given as follows: $\mu_r = 0.4914, \mu_g = 0.4824, \mu_b =  0.4467$; $\sigma_r = 0.2471, \sigma_g = 0.2435, \sigma_b = 0.2616$. Normalization of each channel pixel is done by subtracting the mean value in the corresponding channel and then dividing by the standard deviation of the color channel.

\subsection{Additional Results}
\begin{figure}[!httb] 
\centering

\includegraphics[width=0.48\textwidth]{figures/label_half.png}\\
\subfigure[\textsc{PGD}]{\includegraphics[width=0.23\textwidth]{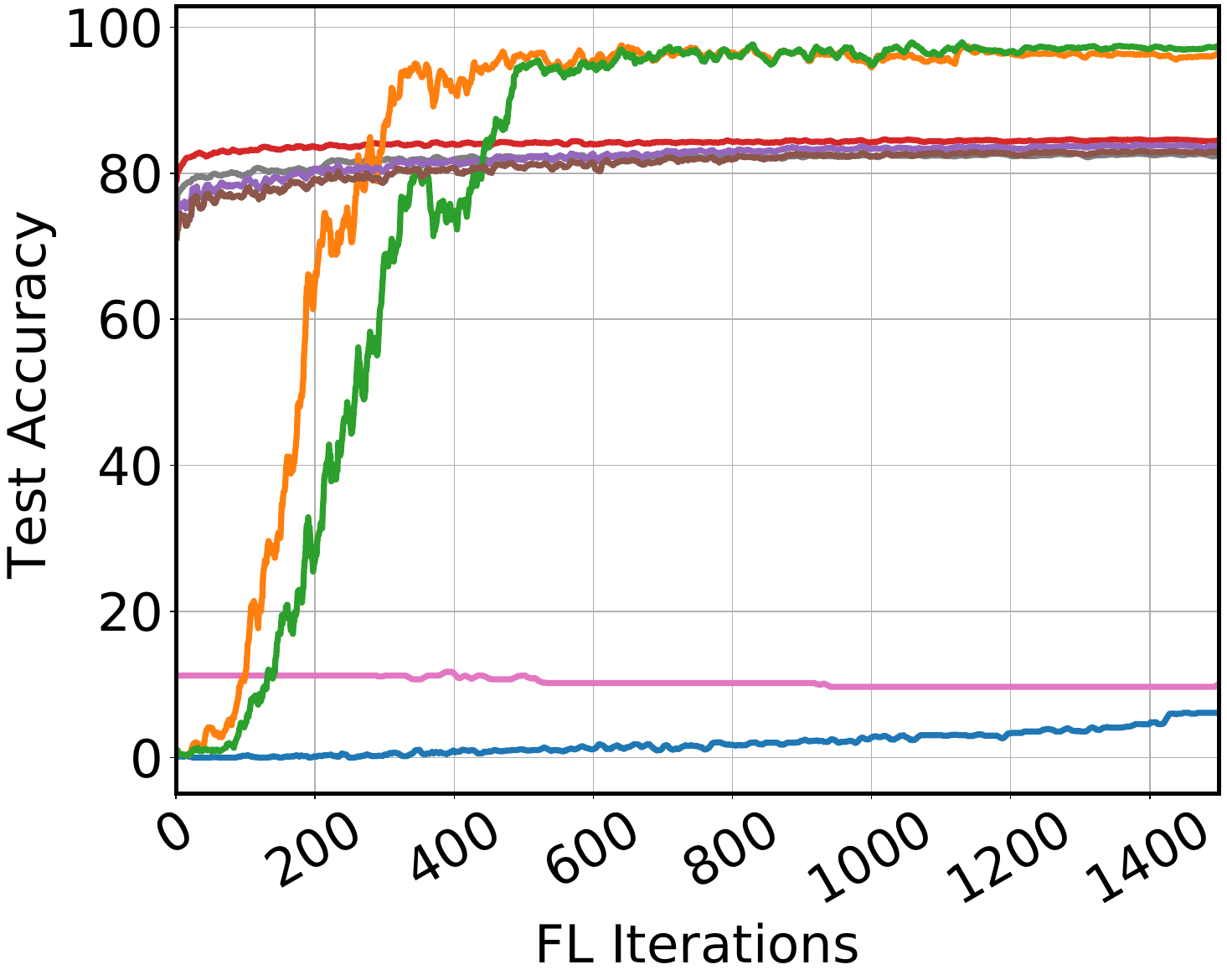}} \subfigure[PGD with replacement]{\includegraphics[width=0.23\textwidth]{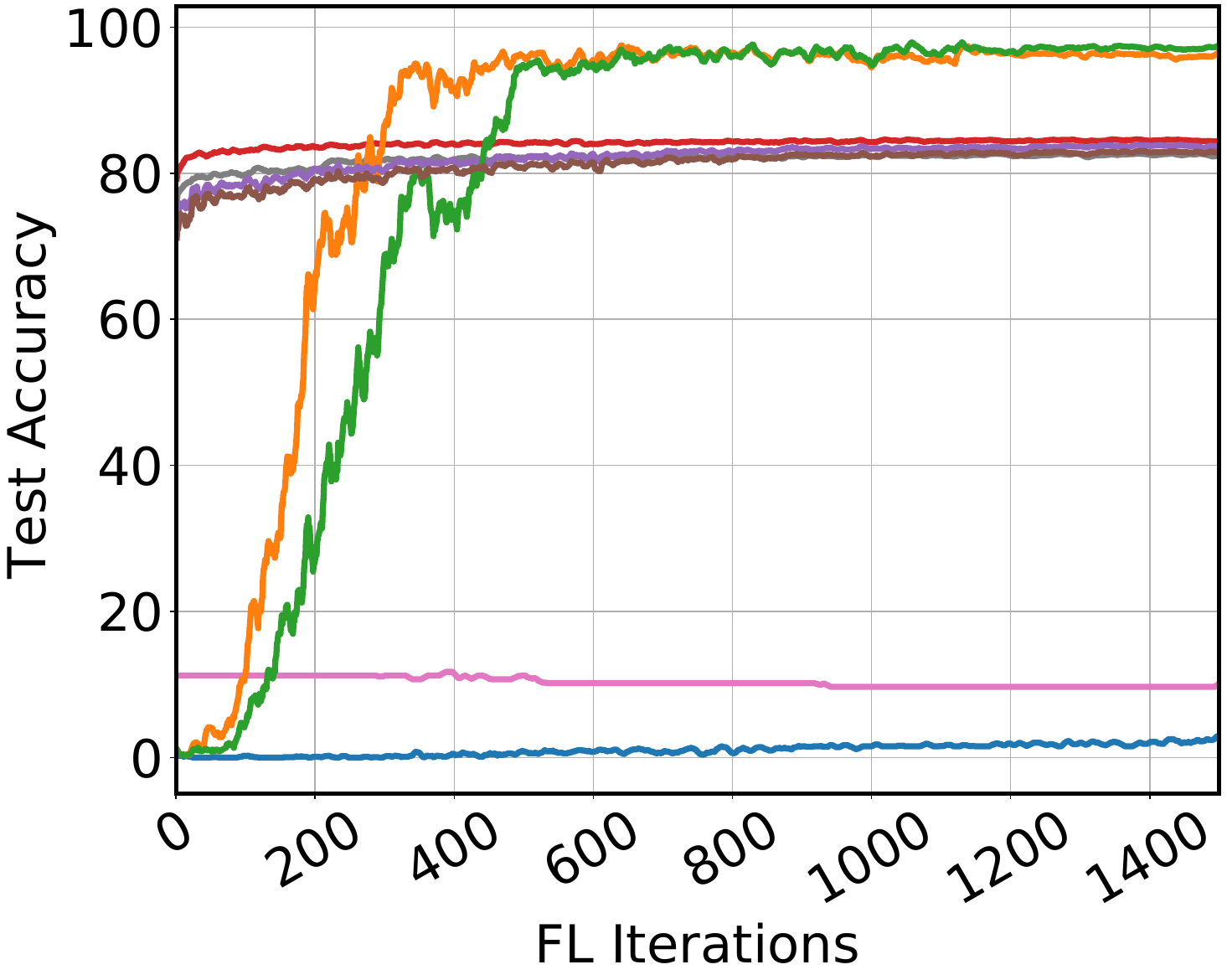}}
\caption{Performance comparison of \ouralgo\ with Sparsefed \cite{panda2022sparsefed} under PGD with/without model replacement for CIFAR-10 Trigger Patch.\protect \footnotemark}
\label{fig:against-attacks_trigger}
\vspace{5mm}
\end{figure}

Figure \ref{fig:against-attacks_trigger} shows the \textit{MA} and \textit{ASR} for two attacks (PGD with and without model replacement) on the CIFAR-10 Trigger Patch dataset. We can observe that in \ouralgo\  ASR, goes lower than 5\%, as learning proceeds. On the contrary, Sparsefed \cite{panda2022sparsefed}, RFA \cite{pillutla2019robust}, and NDC \cite{sun2019can} attain a very high ASR, thus failing to defend against these attacks. It is to be noted that the MA is not at all hampered in our defense while defending against backdoors.
\footnotetext{"MA RFA" has overlapped with "MA NDC-adaptive" and "MA Sparsefed"  due to proximate values.}

\begin{figure}[httb] 
\subfigure[Poison points detected over FL iterations]{\includegraphics[width=0.23\textwidth]{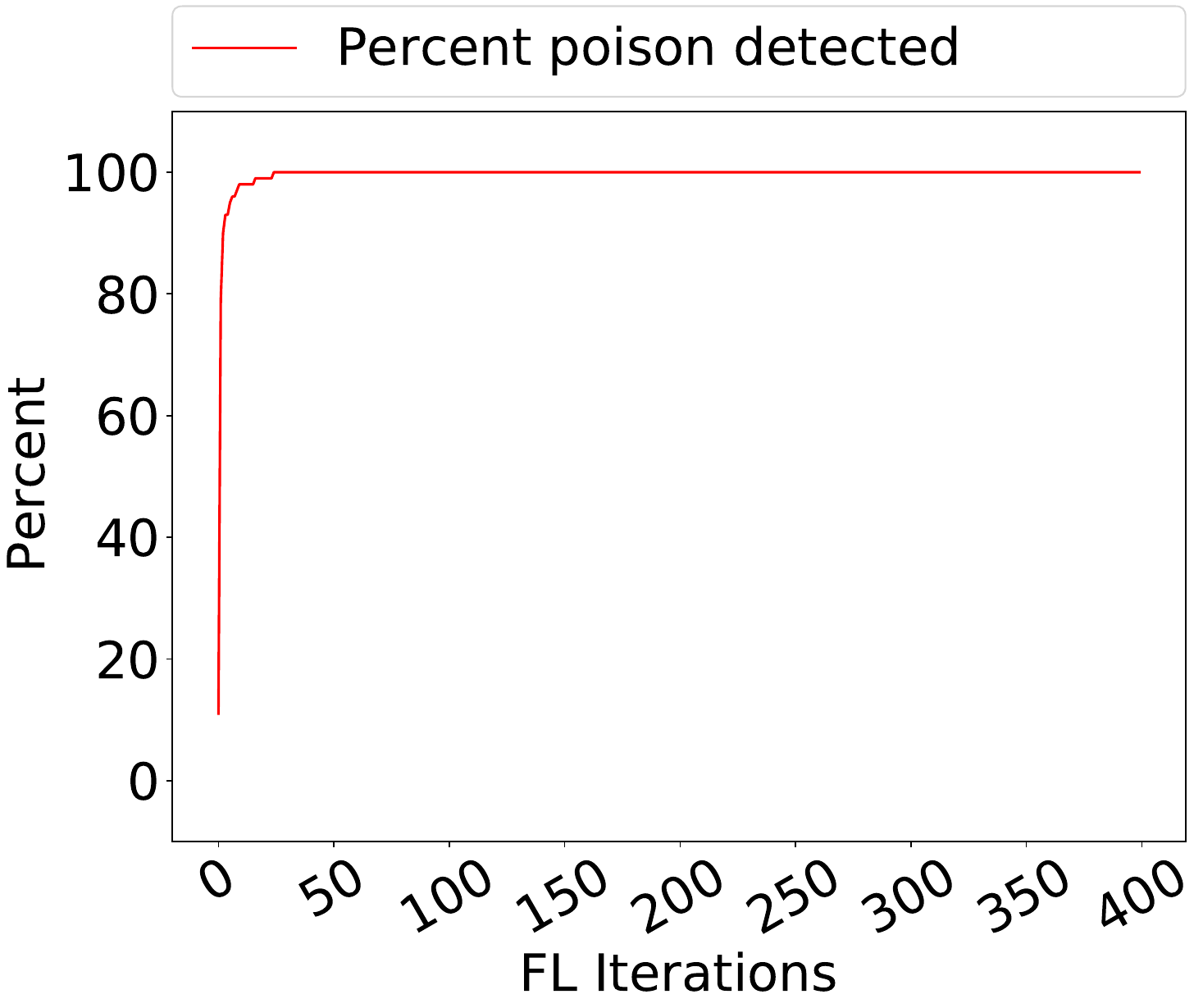}} \hfill
\subfigure[Client Importance difference between attacker and other honest clients]{\includegraphics[width=0.24\textwidth]{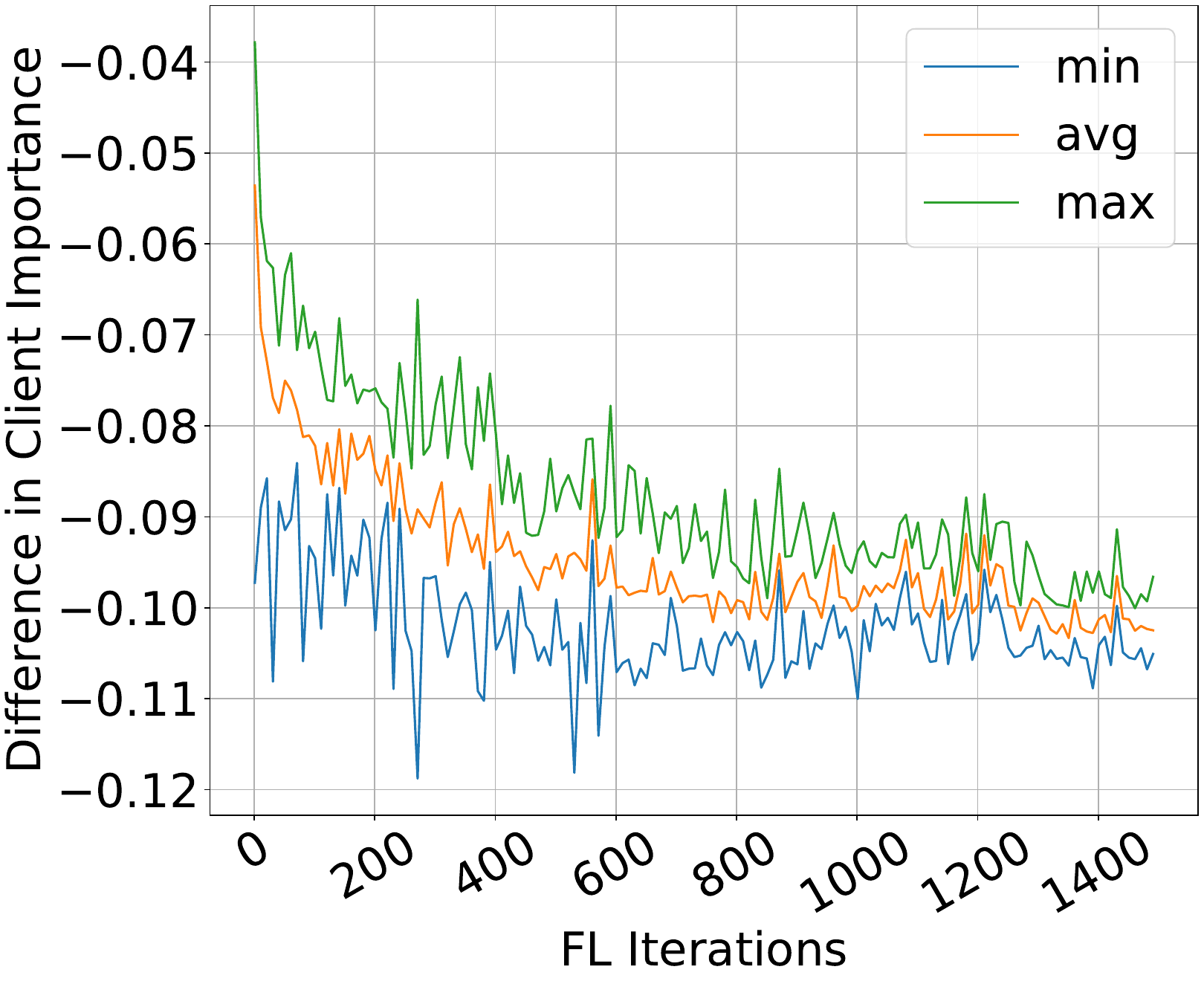}}
\caption{(a) Percent of detected poison points in $D_d$ showing the effectiveness of $\psi$. (b) Analysis of client importance showing the effectiveness of $\theta$ under PGD with model replacement attack for CIFAR-10 Trigger Patch.}
\label{fig:theta_gamma_fig_trigger}
\vspace{6mm}
\end{figure}

We now study the effectiveness of different components of \ouralgo{}. The effectiveness of PDD network parameterized
by $\psi$ is shown in Figure \ref{fig:theta_gamma_fig_trigger}-a. We can observe in the first FL round, \ouralgo{} is able to detect only few poison points (11 out of 100 poison points for CIFAR-10 Trigger Patch). The cost function that is designed to correct the ordering of datapoints such that the PDD gradually detects all the poisoned
points, helps in moving $\psi$ in correct direction. We can see that the number of actual poison points detected are increasing with increasing FL rounds. 

The effectiveness of $\theta$ is shown in Figure \ref{fig:theta_gamma_fig_trigger}-b. We have shown the minimum, maximum and average values of the client importance difference between attacker and the client models in every $10^{th}$ round. We can see a sloping curve that shows that the attacker is given a lower weightage compared to other clients. This leads to an appropriate updation of the global model leading to reduction in ASR.

We also experimented by varying known clean examples ($D_{clean}$) and actual fraction of poisoned points (here we have set $\beta$ the same as the actual fraction of poisoned points) as shown in Table \ref{tab:diff_beta_D_clean} for CIFAR-10 Trigger Patch. We see that \ouralgo{} can work
with as few as 5 poisoned examples ($\beta$ = 0.05). Same study on
CIFAR-10 Southwest is shown in Table \ref{tab:diff_beta_southwest}. Results shows that \ouralgo{} is invariant towards
the different fraction of $D_{clean}$ and $\beta$.

\begin{table}[!httb]
\caption{Sensitivity of \ouralgo{} with different actual percentages of $D_{clean}$ and actual fraction of poisoned points in $D_d$ ($\beta$ is also set same) under PGD with replacement for CIFAR-10 Trigger Patch. 
}
\label{tab:diff_beta_D_clean}
\begin{center}
\begin{small}
\begin{tabular}{p{25 mm}ccc}
\hline\
\textbf{Experiments} & \textbf{Values} &\multicolumn{1}{c}{\begin{tabular}[c]{@{}l@{}}\textbf{MA (\%)} \end{tabular}} & \multicolumn{1}{c}{\begin{tabular}[c]{@{}l@{}}\textbf{ASR (\%)} \end{tabular}}   \\\hhline{====}

\multirow{4}{*}{\begin{tabular}[c]{@{}l@{}}Actual
 percentage\\ of $D_{clean}$ 
\end{tabular}}
    &5\%    & 84.47   &2.04 \\
    &10\%    &84.45   &2.04
      \\
    &15\%    &84.35   &2.55
      \\
    &20\%   &84.47    &2.04
  \\\hline
\multirow{6}{*}{\begin{tabular}[c]{@{}l@{}} Actual fraction \\
of poisoned points
\end{tabular}}
    &0.01    &84.38    &2.93
    \\
    &0.05    &84.41    &2.80
    \\
    &0.1    &84.43    &2.69
    \\
    &0.2    &84.47    &2.04
    \\
    &0.3    &84.50    &1.53
    \\
    &0.5    &84.58    &0.51
  \\\hline

\end{tabular}
\end{small}
\end{center}
\end{table}

\begin{table}[httb]

\caption{Sensitivity of \ouralgo{} with different actual fractions of poisoned examples in $D_d$ ($\beta$ is also set same) under PGD with a replacement for CIFAR-10 Southwest. 
}
\label{tab:diff_beta_southwest}
\begin{center}
\begin{small}
\begin{tabular}{ccc}
\hline\
\multirow{2}{*}{\begin{tabular}[c]{@{}l@{}} \textbf{Actual fraction} \\
\textbf{of poisoned points}
\end{tabular}} &\multirow{2}{*}{\begin{tabular}[c]{@{}l@{}}\textbf{MA (\%)} \end{tabular}} & \multirow{2}{*}{\begin{tabular}[c]{@{}l@{}}\textbf{ASR (\%)} \end{tabular}}   \\\\\hhline{===}


    0.01    &83.62   &20.07    \\
    0.05    &83.85   &18.74     \\
    0.1     &84.16   &16.21     \\
    0.2     &84.49   &15.30     \\
    0.3     &84.54   &5.10     \\
    0.5     &84.67   &3.06     \\\hline

\end{tabular}
\end{small}
\end{center}
\end{table}

\subsection{Final Algorithm}

Algorithm \ref{algo:learning_supple} describes our \ouralgo\ algorithm in detail.

\begin{algorithm}[!t]
  \SetAlgoLined

  \caption{\ouralgo}
  \label{algo:learning_supple}
    	\textbf{Input:  }\\
        {\hspace{4mm} $ \phi$: List of client model parameters \\  
        \hspace{4mm} $D_d$: Defense dataset with both clean and poisoned samples \\
        \hspace{4mm} $D_{clean}$: subset of $D_d$ that are known to be clean.\\
        \hspace{4mm}  $\beta$: fraction of poisoned points to be detected from $D_d$\\
        \hspace{4mm} $T$: FL rounds\\
        } 
            \textbf{Initialize:} \\ 
            
            \hspace{4mm} $\psi^0$ consistent with $D_{clean}$ \Comment{Eq \ref{eq:psi-consistent}}\\
            \hspace{4mm} $\theta^0 \leftarrow \cN(0,1)$ \Comment{Sampling from a gaussian }\\

	    \For {$t=1$ {\bfseries to} $T$ }{
		    
		    Calculate $\gamma^t_{(x_i,y_i) \in D_d}(x_i,y_i, \psi^{t-1} )$
            
             \textbf{Partition $D_d$ into $D^t_{dc}$ and $D^t_{dp}$}	       
	        
	         Sort $\gamma^{t}_i, i\in D_d$ in decreasing order of magnitude.
	       
	         $D^t_{dp}$: High scoring $\beta$ images considered as poisoned,
	        
	         while the remaining as clean $(D^t_{dc})$
		
	         \textbf{Calculate features for Client Importance model}
    		
            
             $ \bar{L}^t_{dc}(\phi_j^t)=  \frac{1}{|D_{dc}^t|} \sum_{(x_i,y_i) \in D_{dc}^t} - log (f_{y_i}(x_i,\phi_j^t))$ 
            
             $ \bar{L}^t_{dp}(\phi_j^t)=  \frac{1}{|D_{dp}^t|} \sum_{(x_i,y_i) \in D_{dp}^t} - log (f_{y_i}(x_i,\phi_j^t))$

	        $dist^t(\phi_j^t) = \left \|\bar{\phi}^{t-1} - \phi^t_j  \right \|_2 $ 
	        
	         $s^t(\phi_j^t) = [\bar{L}^t_{dc}(\phi_j^t), \bar{L}^t_{dp}(\phi_j^t), dist^t(\phi_j^t) ]$
	        
	       
	         CI model: $\mathcal{C}^t(\phi_j^t;\theta^{t-1}) = {\theta^{t-1}}^T s^t_j ; \forall j={1,...,M}$
	       
	         Normalise $\mathcal{C}^t(\phi_j^t;\theta^{t-1}) = \frac{ReLu(\mathcal{C}^t(\phi_j^t;\theta^{t-1}))}{\sum_j ReLu(\mathcal{C}^t(\phi_j^t;\theta^{t-1}))} $.
    		
    	\textbf{Calculate the global model} 
    		
    	$\bar{\phi}^t(\theta^{t-1}) \leftarrow \sum_{j=1}^M(\mathcal{C}^t(\phi^t_j,\theta^{t-1})  * \phi^t_j) $

            \textbf{Compute loss using the updated global model} 
           
            $l^t_{c}((x_i,y_i);\bar{\phi}^t) = - log(f_{y_i}(x_i;\bar{\phi}^t)) ; \forall (x_i,y_i) \in D_d$
              	
    	 $l^t_{p}((x_i,y_i);\bar{\phi}^t)    \hspace{-1.1mm}=\hspace{-1.1mm} - log(1-f_{y_i}(x_i;\bar{\phi}^t)); \forall (x_i,y_i) \hspace{-1mm} \in \hspace{-1mm} D_d$

    	 $\mathcal{L}^t_{\theta}\hspace{-1mm}= \hspace{-0.84mm}\sum_{(x_i,y_i) \in D^t_{dc}} l^t_{c}(x_i,y_i) \hspace{-0.84mm}+\hspace{-0.84mm} \sum_{(x_i,y_i) \in D^t_{dp}} l^t_{p}(x_i,y_i)$

    	 \textbf{Update client importance model parameter $\theta$}
    		
    	 $\theta^t = \theta^{t-1} - \alpha \nabla_{\theta} \mathcal{L}_{\theta}^t$

	     \textbf{Calculate the cost function}\\
	     $V(\psi^{t-1}|D_d,\bar{\phi}^t)
	    \hspace{-0.8mm}=\hspace{-1mm} \sum_i\hspace{-1mm} \gamma_i^t(\psi^{t-1})(l_{p}^t(x_i;\bar{\phi}^t) \hspace{-1mm}- l_{c}^t(x_i;\bar{\phi}^t) $
	        
	     \textbf{Update PDD  parameter $\psi$} \\
    	 $\psi^t = \psi^{t-1} - \eta \nabla_{\psi} V(\psi^{t-1}|D_d,\bar{\phi}^t)$\\

            }
     	\textbf{Output:  } Global Model $\bar{\phi}$
\end{algorithm}

\end{document}